\documentclass{article}
\usepackage{spconf,amsmath,graphicx,hyperref}
\usepackage{algpseudocode}
\usepackage[english]{babel}
\usepackage{mathptmx}
\usepackage{booktabs}
\usepackage[parfill]{parskip}
\usepackage{enumitem}

\usepackage{cite}
\usepackage{mathtools}
\usepackage{amsmath}
\usepackage{amssymb}
\usepackage{amsthm}
\usepackage{graphicx} 
\usepackage{hyperref}
\usepackage{comment}
\usepackage{subcaption}
\usepackage{xcolor}
\usepackage{nccmath}

\usepackage{fancyhdr}
\usepackage{array}    
\usepackage{tabularx}
\usepackage{upgreek}

\setlist{leftmargin=*, itemsep=0pt, topsep=2pt, parsep=0pt, partopsep=0pt}

\usepackage[font=small,labelfont=bf]{caption}
\makeatletter
\g@addto@macro\normalsize{%
  \setlength\abovedisplayskip{4pt}
  \setlength\belowdisplayskip{4pt}
  \setlength\abovedisplayshortskip{2pt}
  \setlength\belowdisplayshortskip{2pt}
}
\makeatother
\usepackage[final]{microtype}

\makeatletter
\algrenewcommand\algorithmicindent{0.8em}
\makeatother

\newtheorem{theorem}{Theorem}
\newtheorem{lemma}{Lemma}
\theoremstyle{remark}
\newtheorem{remark}{Remark}
\newcommand{\R}{\mathbb{R}}

\newenvironment{proofsketch}{%
  \proof}{\endproof}

\usepackage{algorithm}

\makeatletter
\renewcommand\section{\@startsection {section}{1}{\z@}
  {-6pt plus -2pt minus -2pt}
  {3pt plus 1pt minus 1pt}%
  {\normalfont\large\bfseries}}   

\renewcommand\subsection{\@startsection{subsection}{2}{\z@}%
  {-4pt plus -2pt minus -2pt}%
  {2pt plus 1pt minus 1pt}%
  {\normalfont\normalsize\bfseries}}
\makeatother

\renewenvironment{thebibliography}[1]{%
  \begin{oldthebibliography}{#1}%
    \setlength{\parskip}{0pt}%
    \setlength{\itemsep}{0pt plus 0.3ex}%
}{\end{oldthebibliography}}

\title{Accelerated Sinkhorn Algorithms for Partial Optimal Transport}

\name{\thanks{$^*$ The authors assert joint first authorship}
Nghia Thu Truong$^{*\ddag}$, \qquad
Qui Phu Pham$^{*\S}$, \qquad
Quang Nguyen$^{*\dagger\clubsuit}$, \qquad
Dung Luong$^{\#}$, \qquad
Mai Tran$^{\star}$}

\address{
$^{\ddag}$University of Maryland, College Park, USA\\
$^{\S}$University of California, Irvine, USA\\
$^{\dagger}$University of Information Technology, Ho Chi Minh City, Vietnam\\
$^{\clubsuit}$Vietnam National University, Ho Chi Minh City, Vietnam\\
$^{\#}$VietDynamic, HCMC, Vietnam\\
$^{\star}$Binh Duong University, Binh Duong, Vietnam
}

\begin{document}

\maketitle
\begin{abstract}
Partial Optimal Transport (POT) addresses the problem of transporting only a fraction of the total mass between two distributions, making it suitable when marginals have unequal size or contain outliers. While Sinkhorn-based methods are widely used, their complexity bounds for POT remain suboptimal and can limit scalability. We introduce Accelerated Sinkhorn for POT (ASPOT), which integrates alternating minimization with Nesterov-style acceleration in the POT setting, yielding a complexity of $\mathcal{O}(n^{7/3}\varepsilon^{-5/3})$. We also show that an informed choice of the entropic parameter $\gamma$ improves rates for the classical Sinkhorn method. Experiments on real-world applications validate our theories and demonstrate the favorable performance of our proposed method. Our code is open-source at the link \href{https://github.com/qpp112/Accelerated-Sinkhorn-Algorithms-for-Partial-Optimal-Transport.git}{github}.


\end{abstract}


\section{Introduction}
\label{sec:intro}
Optimal Transport (OT) has been a core tool to compare probability distributions, with broad impact from economics to machine learning \cite{dvurechensky2018computational,pmlr-v84-genevay18a,robust_uot,JMLR_UOT}. Its success stems from efficient computational methods \cite{doi:10.1137/141000439,cuturi2013sinkhorndistanceslightspeedcomputation,dvurechensky2018computational,lin2022efficiency} and the role of entropic regularization, which improves statistical properties and enables seamless integration into deep learning frameworks \cite{balaji2020robustoptimaltransportapplications,cuturi2013sinkhorndistanceslightspeedcomputation,pmlr-v84-genevay18a,mena2019statisticalboundsentropicoptimal,bonneel2023survey,chen2023plotpromptlearningoptimal}. 
Classical OT, however, requires equal mass and is sensitive to outliers \cite{raghvendra2024new}.
POT addresses these issues by relaxing mass constraints, making it more robust in practical tasks such as image analysis, few-shot learning and distributional learning \cite{DBLP:journals/corr/abs-2404-03446,DBLP:conf/eccv/ZhengHTX24,zhang2024pot,le2022multimarginal,nguyen2024partial,NEURIPS2020_1e6e25d9}. However, existing computational approaches remain unsatisfactory. A practical path to POT is to reduce it to a balanced OT instance by adding dummy nodes and then apply Sinkhorn‐type iterations \cite{NEURIPS2020_1e6e25d9}. While straightforward, naive variants can be infeasible \cite{le2022multimarginal}. A feasible formulation with guarantees was established in \cite{nguyen2024partial}, yielding a Sinkhorn complexity of $\mathcal{O}\left(n^2/\varepsilon^4\right)$ and proposing APDAGD with $\mathcal{O}\left(n^{2.5}/\varepsilon\right)$. However, applied researchers note that tuning the Sinkhorn parameter often yields stronger performance, while APDAGD’s extra $\mathcal{O}(n^{0.5})$ factor limits its scalability. Thus, our central questions are:
\begin{enumerate}[leftmargin=*]
    \item Is there a first-order method for entropic POT with a better $\varepsilon$-dependence than feasible Sinkhorn and a better $n$-dependence than APDAGD?
    \item How does varying the regularization parameter $\gamma$ affect the feasible Sinkhorn for POT, and can parameter tuning improve both theory and practice?
\end{enumerate}
We summarize our contributions as follow:
\begin{enumerate}[leftmargin=*]
    \item We propose ASPOT, the first accelerated Sinkhorn for POT, with complexity $\mathcal{O}(n^{7/3}\varepsilon^{-5/3})$, yielding a better $n$–$\varepsilon$ trade-off and bridging to near-linear OT algorithms.
    \item We analyze the classic Sinkhorn, showing how the careful choice of $\gamma$ improves both theory and practice, achieving a better complexity $\mathcal{O}\left(n^2\|C\|^2_{\infty}/\varepsilon^{\frac{3p+1}{p}}\right)$. Here $p\in[1,\infty)$ is a tunable analysis parameter. We write $o(1)$ to denote a quantity that can be made arbitrarily small by taking $p$ sufficiently large.
    \item Through experiments on color transfer and point cloud registration, we show that ASPOT converges faster and produces higher-quality solutions than existing baselines, and that tuning $\gamma$ further boosts Sinkhorn’s performance.
\end{enumerate}

\begin{table}[h!]
\centering
\small
\begin{tabular}{l c}
\toprule
\textbf{Algorithm} & \textbf{Complexity} \\
\midrule
Feasible Sinkhorn \cite{nguyen2024partial} & $\mathcal{O}\!\left(n^2\|C\|_\infty^2/\varepsilon^4\right)$\\
APDAGD \cite{nguyen2024partial} & $\mathcal{O}\!\left(n^{2.5}\|C\|_\infty/\varepsilon\right)$ \\
\textbf{ASPOT (This paper)} & $\mathcal{O}\!\left(n^{7/3}\|C\|_\infty^{4/3}/\varepsilon^{5/3}\right)$ \\
\textbf{Tuned Sinkhorn (This paper)} & $\mathcal{O}\!\left(n^2\|C\|_\infty^2/\varepsilon^{3+o(1)}\right)$\\
\bottomrule
\end{tabular}
\caption{Complexity of POT algorithms}
\label{table}
\end{table}

\section{Preliminaries}
\label{sec:preliminaries}

\subsection{Problem formulation}
\textbf{Notation:} Vectors are bold lowercase, matrices bold uppercase, and $\mathbf{1}_n$ is the all-ones vector in $\mathbb{R}^n$. For any vector or matrix $X$, $X^\top$ denotes transpose, $\langle A,B\rangle=\sum_{i,j}A_{ij}B_{ij}$, and $H(x)=-\langle x,\log x\rangle$. We use $\|x\|_1=\sum_i|x_i|$, $\|A\|_{1\to2}=\max_{\|x\|_1=1}\|Ax\|_2$, $\rho(a,b)=b-a+a\log(a/b)$ for $a,b>0$, and $\operatorname{diag}(x)$ for the diagonal matrix with $x$ on its diagonal. Dual variables are $z=(u,v,w)$ with $\nabla$ their derivatives. Algorithmic complexity is denoted $\mathcal{O}(\cdot)$.

Firstly, recall the POT problem. Given two discrete distributions $r,c \in \mathbb{R}^n_+$ with possibly different masses and a transport budget \( s \leq \min \{ \|r\|_1, \|c\|_1 \} \), the goal is to find a transport plan $X \in \mathbb{R}^{n\times n}_+$ minimizing the cost  
\begin{align}
 \text{POT}(r,c,s) = \min_{X \in \mathcal{U}(r,c,s)} \langle C, X \rangle 
\end{align}
where $C \in \mathbb{R}^{n\times n}_+$ is the cost matrix and the feasible set is $
\mathcal{U}(r,c,s) = \{ X \geq 0 : X \mathbf{1}_n \leq r,\, X^\top \mathbf{1}_n \leq c,\, \mathbf{1}_n^\top X \mathbf{1}_n = s \}.$ 

Following the success of entropic OT, entropic regularization leads to $  \text{min}_{X \in \mathcal{U}(r,c,s)} \langle C, X \rangle - \gamma H(X)$, where $\gamma > 0$ is the regularization parameter. The corresponding dual \cite{nguyen2024partial} introduces potentials $u,v \in \mathbb{R}^n$ and $w \in \mathbb{R}$, giving
\begin{align}
\label{dual}
\notag
 \min_{z} \varphi(z) = \|B(z)\|_1 &+ \sum_{i=1}^{n}\exp(u_i) + \sum_{j=1}^{n}\exp(v_j) \\
 &- \langle u, r \rangle - \langle v, c \rangle - w s ,
\end{align}
where $B(z) := m D_1 K D_2$, with $D_1 = \mathrm{diag}(e^{u_i})$, $D_2 = \mathrm{diag}(e^{v_j})$, $K = e^{-C/\gamma}$, and $m = e^w$.

Next, we define the standard notion of $\varepsilon$-approximation:

\textbf{Definition 1.} For $\varepsilon > 0$, a transport plan $X$ is an $\varepsilon$-approximation of POT($r,c,s$) if
$
X \in \mathcal{U}(r,c,s), \langle C, X \rangle \leq \langle C, X^\star \rangle + \varepsilon,
$ where $X^\star$ is an optimal transport plan of the POT problem.

\section{Algorithm Development}
Both proposed methods share a two-stage structure where in the first stage, we design a solver tailored to the entropic-regularized POT problem.
In the second stage, we invoke the \textsc{Approximating POT} procedure (Algorithm~\ref{AcceleratedSinkhorn}, Lines~1–8) together with the \textsc{ROUND-POT} routine, which ensures an $\varepsilon$-approximate solution that remains feasible.
Coupled with our ASPOT and Tuned Sinkhorn variants, this framework achieves improved complexities of $\mathcal{O}(n^{7/3}\varepsilon^{-5/3})$ and $\mathcal{O}(n^2/\varepsilon^{(3p+1)/p})$, respectively.
A full comparison of POT complexities for existing and proposed methods is summarized in Table~\ref{table}.

\begin{algorithm}[t]
\caption{\textbf{Greenkhorn Step}$(u,v,w)$}
\label{alg:greenkhorn}
\begin{algorithmic}[1]
\If{$t \bmod 3 = 0$} 
    \State $u \gets u + \log r - \log\!\big(r(B(u,v,w)) + e^u\big)$
\ElsIf{$t \bmod 3 = 1$} 
    \State $v \gets v + \log c - \log\!\big(c(B(u,v,w)) + e^v\big)$
\Else 
    \State $w \gets w + \log s - \log \|B(u,v,w)\|_1$
\EndIf
\State \Return $(u,v,w)$
\end{algorithmic}
\end{algorithm}
\begin{theorem}
\label{overall complexity}
Algorithm \ref{AcceleratedSinkhorn} returns an $\varepsilon$-approximate transport plan in $\tilde{O}\left(\frac{n^{7/3}\|C\|_{\infty}^{4/3}(\log(n))^{1/3}}{\varepsilon^{5/3}}\right)$ arithmetic operations.
\end{theorem}
\begin{remark}
Theorem \ref{bound iterations for Accel} establishes the iteration complexity required for Algorithm~\ref{AcceleratedSinkhorn} to reach an $\varepsilon$-approximate solution. 
 Since each iteration costs only $\tilde{O}(n^2)$ arithmetic operations, this yields the overall complexity stated in Theorem~\ref{overall complexity}.
\end{remark}

\begin{algorithm}[t]
\caption{Accelerated Sinkhorn POT}
\label{AcceleratedSinkhorn}
\begin{algorithmic}[1]
\Require Marginals $r,c$, cost matrix $C$, transport mass $s$, accuracy $\varepsilon$
\State Set $\gamma = \varepsilon /(4\log n)$,\quad $\tilde{\varepsilon}=\varepsilon /(8\|C\|_{\max})$
\If{$\|r\|_1>1$} \State $\tilde{\varepsilon}\gets\min\{\tilde{\varepsilon},\, \tfrac{8(\|r\|_1-s)}{\|r\|_1-1}\}$ \EndIf
\If{$\|c\|_1>1$} \State $\tilde{\varepsilon}\gets\min\{\tilde{\varepsilon},\, \tfrac{8(\|c\|_1-s)}{\|c\|_1-1}\}$ \EndIf
\State $\tilde{r}=(1-\tfrac{\tilde{\varepsilon}}{8})r+\tfrac{\tilde{\varepsilon}}{8n}\mathbf{1}_n$, \quad
       $\tilde{c}=(1-\tfrac{\tilde{\varepsilon}}{8})c+\tfrac{\tilde{\varepsilon}}{8n}\mathbf{1}_n$
\State Initialize $t=0$, $\theta_0=1$, 
$\tilde{z}^0=\check{z}^0=\mathbf{0}_{n+1}$.
\While{$E_t>\tilde{\varepsilon}$}
  \State $\bar{z}^t \gets (1-\theta_t)\check{z}^t+\theta_t \tilde{z}^t$
  \State $\tilde{z}^{t+1}\!\gets\!\bar{z}^t-\frac{\gamma}{3(\|r\|_1+\|c\|_1-s)\theta_t}\nabla\varphi(\bar{z}^t)$
  \State $\grave{z}^t \gets \bar{z}^t + \theta_t\,(\tilde{z}^{t+1}-\tilde{z}^t)$
  \State $\hat{z}^t \gets \textsc{GreenkhornStep}(\grave{z}^t)$
  \State $z^t \gets \arg\min\{\varphi(z): z\in\{\check{z}^t,\hat{z}^t\}\}$
  \State $\check{z}^{t+1} \gets \textsc{GreenkhornStep}(z^t)$
  \State $\theta_{t+1}\gets \frac{\theta_t\big(\sqrt{\theta_t^2+4}-\theta_t\big)}{2}$,\quad $t\gets t+1$
\EndWhile
\State \textbf{Form primal:} $\widetilde{X} \gets B(z^t)$ with
       $B_{ij}(z)=\exp\!\big(-C_{ij}/\gamma+u_i+v_j+w\big)$
 \State \textbf{Round} $\widetilde{X}$ to $\bar{X}$ using \textsc{ROUND-POT}\cite{nguyen2024partial}
\State \textbf{Output:} $\bar{X}$
\end{algorithmic}
\end{algorithm}

\subsection{Acceleratd Sinkorn POT Algorithm}
Algorithm~\ref{AcceleratedSinkhorn} is designed to combine two ideas: the momentum mechanism of Nesterov and the fast per-iteration progress of Greenkhorn in low-precision regimes.  
A naive combination, such as the Alternating Accelerated Method (AAM) \cite{altschuler2017near}, leads to an additional $\mathcal{O}\left(n^{0.5}\right)$ multiplicative factor in the complexity, which makes it impractical for large-scale problems in practice.  
Instead, we let the Greenkhorn update serve directly as the coupling step within Nesterov’s framework in line 12. After each extrapolation, the algorithm applies the most effective coordinate correction, stabilizing the momentum and ensuring that progress concentrates on the largest residual (line 14). To monitor this, we use a function measuring the error per iteration:
\begin{align}
E(t) = \big|\mathbf{1}_n^\top B^t \mathbf{1}_n - s\big|
       + \|B^t \mathbf{1}_n + \tilde{p} - r\|_1
       + \|(B^t)^\top \mathbf{1}_n + \tilde{q} - c\|_1, \notag
\end{align}
Equally important is the descent of the dual objective \ref{dual}. The Greenkhorn updates in lines~14 and~16 enforce  
$\varphi(\grave{z}^t)\!\ge\!\varphi(\hat{z}^t)$ and $\varphi(z^t)\!\ge\!\varphi(\check{z}^{t+1})$, ensuring steady decrease for \ref{dual}.

\begin{figure}[t]
    \vspace{-4pt}
    \centering
\includegraphics[width=1.0\linewidth]
    {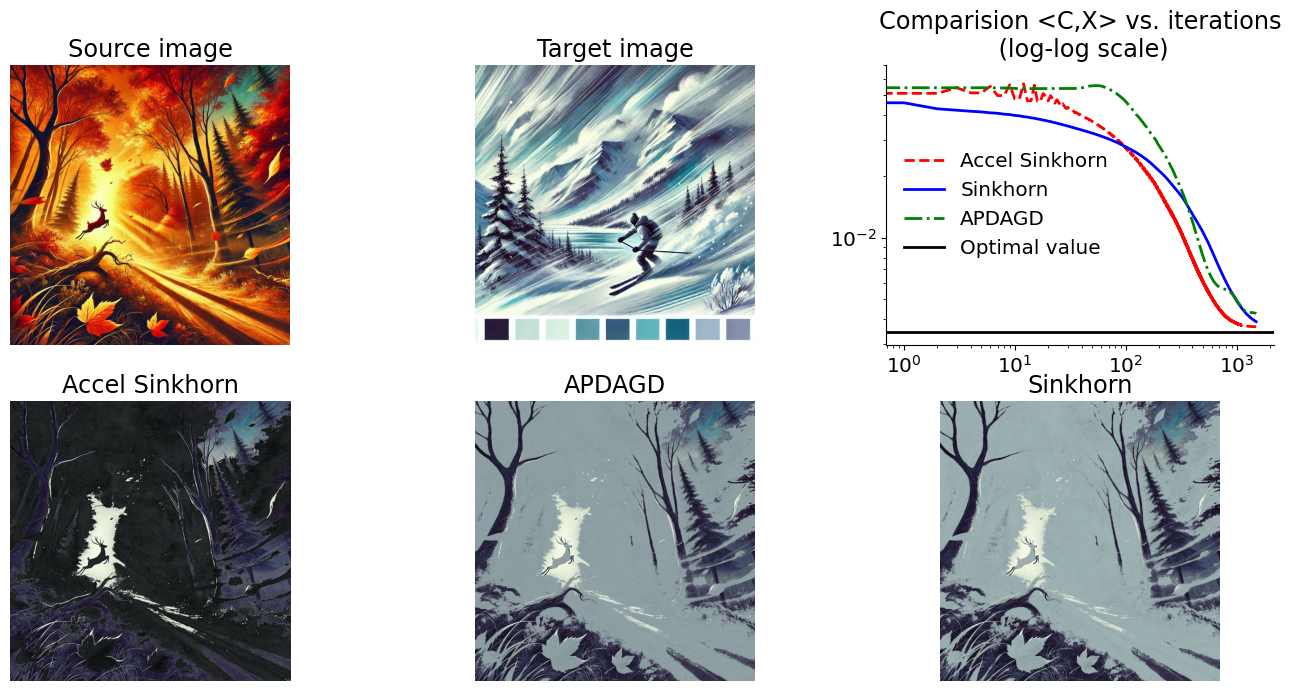}
    \caption{Color Transfer Experiment}
        \vspace{-6pt}
    \label{fig:AI picture}
\end{figure}

\subsection{Tuned Sinkhorn Technique}
Motivated by recent insights in the OT literature \cite{kemertas2025efficient}, we revisit the role of the regularization parameter $\gamma$ in the Sinkhorn algorithm for POT \cite{nguyen2024partial}. By scaling $\gamma$ relative to the minimal entropy term $H_{\min}(r,c) = \min\{H(r),H(c)\}$ and the accuracy parameter $\varepsilon$, the classical Sinkhorn method attains an improved complexity bound of $\mathcal{O}\left(n^2H_{\text{min}}(r,c)^{1/p}\|C\|^2_{\infty}/\varepsilon^{\frac{3p+1}{p}}\right)$. This refined analysis clarifies the performance of Sinkhorn in practice and extends its theoretical guarantees to the partial transport setting.
\begin{theorem}
\label{tunedSinkhorn}
The feasible Sinkhorn algorithm \cite{nguyen2024partial} for POT with the chosen regularization $\gamma = \sqrt[p]{2\varepsilon/49H_{\text{min}}(r,c)}$ and $p \in [1,\infty)$ returns the $\varepsilon$-approximation in 
\begin{align}
\mathcal{O}\left(n^2H_{\text{min}}(r,c)^{1/p}\|C\|^2_{\infty}/\varepsilon^{3+o(1)}\right)
\end{align}
arithmetic operations.
\end{theorem}
\begin{proofsketch} The error $\langle \bar{X} - X^{\star},C\rangle$ decomposes into rounding, regularization, and the gap between the extended OT solution and the POT optimum. The rounding error is guaranteed by \textsc{ROUND-POT} in Algorithm \ref{AcceleratedSinkhorn} \cite{nguyen2024partial}, contributing $O(\varepsilon')$. The regularization error is controlled by the entropy-regularized OT analysis \cite{kemertas2025efficient}, which also applies to POT since the extra $(n{+}1,n{+}1)$ entries scale with $1/\gamma$ to ensure stability. Finally, Proposition~1 in \cite{NEURIPS2020_1e6e25d9} bounds the remaining gap within $O(\varepsilon')$. Combining yields $\langle \bar{X} - X^{\star},C\rangle \le \varepsilon + \tilde{O}(\varepsilon^2)$. Substituting $\varepsilon' = H_{\min}(r,c)\gamma^p$, $A=\tilde{O}(\|C\|_{\infty}/\varepsilon)$, and $\gamma = \sqrt[p]{2\varepsilon/(49H_{\min}(r,c))}$, together with the $O(n^2)$ per-iteration cost of Sinkhorn, completes the argument.
\end{proofsketch}

\subsection{Theoritical lemmas and theorems}We refer the reader to Section \ref{sec:appendix} for the construction of the matrix $A$ (via slack variables turning POT into a linear programm), complete proofs, extra algorithms, and further experiments. From \cite{nguyen2024partial}, we recall the bound on the dual variables $z^{\star}$: 
\begin{align}
\label{eq:R}
    R := \frac{\|C\|_{\infty}\max(\|r\|_{1},\|c\|_{1})}{\gamma(\max(\|r\|_{1},\|c\|_{1})-s)} - \log\!\Big(\min_{1\le i,j \le n}(r_i,c_j)\Big).
\end{align}

\begin{lemma}
\label{lemma:1}
    Define the Lipschitz constant $L = \frac{\|A\|^2_{1\rightarrow 2}}{\mu_f}$ where $\mu_f = \frac{\gamma}{\|r\|_1+\|c\|_1-s}$.  
    Let $\delta_t = \varphi(\check{z}^t) - \varphi(z^{\star})$, where $z^{\star}$ is an optimal dual solution of the entropic-regularized POT problem.  
    Then we have
    \[
    \delta_{t+1} \leq (1-\theta_t)\delta_t 
    + \tfrac{L}{2}\theta_t^2 \Big(
    \|z^{\star}-\tilde{z}^t\|^2
    - \|z^{\star}-\tilde{z}^{t+1}\|^2
    \Big).
    \]
\end{lemma}

\begin{proofsketch}
The key step to derive $\delta_t$ is to couple the extrapolated iterate $\bar z^t$ with both the new accelerated point $\tilde z^{t+1}$ and the previous $\tilde z^t$, which produces a telescoping quadratic term $\|z^{\star}-\tilde z^t\|^2-\|z^{\star}-\tilde z^{t+1}\|^2$. This coupling makes the quadratic distance terms cancel in a telescoping way, so the error can only get smaller over time. Finally, the Greenkhorn corrections enforce a monotone descent of the dual objective, namely 
$\varphi(\grave z^t)\!\ge\!\varphi(\hat z^t)\!\ge\!\varphi(z^t)\!\ge\!\varphi(\check z^{t+1})$, 
allowing the recurrence to be transferred to the actual iterates. 
\end{proofsketch}

\begin{remark}
  Lemma \ref{lemma:1} allows us to express $\delta_{t+1}$ recursively in terms of $\delta_t$ and a decreasing potential involving $\|z^{\star}\|$. Next, from the definition of $\theta_t$ implies the relation $\tfrac{\theta_{t+1}}{\theta_t}=\sqrt{1-\theta_{t+1}}$. Hence, by iterating this recursion, the error is controlled by the norm of the optimal dual solution. Finally, invoking the uniform bound $\|z^{\star}\| \leq \sqrt{2n}\,R$ leads directly to the quantitative decay rate stated in Lemma \ref{lemma:2}.
\end{remark}

\begin{lemma}
\label{lemma:2}
Let $\{\check{z}^t\}_{t \geq 0}$ be the iterates generated by Algorithm \ref{AcceleratedSinkhorn}, and $z^{\star}$ be an optimal solution of the dual entropic regularized POT problem satisfying $\|z^{\star}\| \leq \sqrt{2n}\,R$. Then we have $\delta_t \leq \frac{4nLR^2}{(t+1)^2}.$
\end{lemma}

\begin{remark}
 
The $O(1/t^2)$ rate controls the dual optimality gap, but the algorithm terminates based on the feasibility error $E(t)$. By relating $E(t)$ to the potential decrease in each iteration, one obtains an iteration bound in terms of $L,R,n$ and the target $\varepsilon'$, we end up with Theorem \ref{bound iterations for Accel}.   
\end{remark}

\begin{theorem}
\label{bound iterations for Accel}
 Let $\left\{\left(z^t\right)\right\}_{t \geq 0}$ be the iterates generated by Algorithm \ref{AcceleratedSinkhorn}. The number of iterations required to reach the stopping criterion $E_t \leq \varepsilon^{\prime}$ satisfies $t \leq 1+\left(\frac{12 \sqrt{14nL} R}{\varepsilon^{\prime}}\right)^{2 / 3}.$
\end{theorem}
\begin{proofsketch}
The proof proceeds by bounding the potential decrease $S=\varphi\left(z^t\right) -\varphi\left(\check{z}^{t+1}\right)$. If the update is on $u$ (or symmetrically $v$), one rewrites $S$ in terms of the marginal constraints and shows that it reduces to a sum of divergences $\rho(r_i, r_i(B(z^t)+\tilde p_i^t)$. If the update is on $w$, a similar calculation using the update rule for $w^{t+1}$ shows that $S=\rho(s,\|B(z^t)\|_1)$. Combining both cases yields a lower bound for $S$ in terms of all marginal violations, namely $S \ge \tfrac{1}{3}\sum_i\rho(\cdot)+\tfrac{1}{3}\sum_j\rho(\cdot)+\tfrac{1}{3}\rho(\cdot)$. Next, Lemma 6 of \cite{altschuler2017near}, together with Cauchy–Schwarz, shows that $S \ge \tfrac{1}{63}E(t)^2$.  
This gives a telescoping inequality on the dual objective $\varphi$.  
Using Lemma~\ref{lemma:2}, we obtain $\sum_{i=j}^t E_i^2 \le 252nLR^2/(j+1)^2$. Since $E_t$ must stay above the stopping threshold $\varepsilon'$ (line 10 in Algorithm~\ref{AcceleratedSinkhorn}), this would lead to a contradiction unless $t$ is bounded. Now choosing $j=t/2$ yields the final rate $t \le 1 + \left(\tfrac{12\sqrt{14nL}R}{\varepsilon'}\right)^{2/3}$.
\end{proofsketch}

\section{Experimental results}
\label{sec:experimental results}

\subsection{Color transfer}
\label{sec:color}
\textbf{Experimental setup.} We follow the setup of \cite{blondel2018smoothsparseoptimaltransport}.  
AI-generated images (Fig.~\ref{fig:AI picture}) are represented in $\mathbb{R}^3$ by RGB values of pixels and quantized to $n=800$ colors using $k$ means, producing histograms $a,b$ as source and target distributions.  
Both are normalized by $\max\{\|a\|_1,\|b\|_1\}$ so total mass $\le1$.  
The cost is $C_{ij}=\|a_i-b_j\|^2/\max(C)$, and the transport budget is $s=0.2\cdot \min(\sum a,\sum b)$.  
We compare ASPOT, feasible Sinkhorn, and APDAGD, each run for 1500 iterations with $\gamma=10^{-3}$ and $\texttt{tol}=10^{-7}$.  
After computing a plan $X$, barycentric projection $\hat a_i=\sum_j X_{ij}b_j/\sum_j X_{ij}$ recolors pixels assigned to centroid $a_i$, transforming the new image.
\textbf{ASPOT results.}  
Fig.~\ref{fig:AI picture} compares the three methods on color transfer.  
The convergence curves show that ASPOT rapidly reduces transport cost, reaching good solutions far earlier than the others. Sinkhorn progresses steadily but much slower, while APDAGD stalls for many iterations before improving. As a result, when ASPOT is already near optimal, the other methods still lag by a gap consistent with ASPOT’s better $\varepsilon$-dependence. The visual results confirm that APDAGD and Sinkhorn yield almost identical outputs, both washed out, lacking detail, with incomplete recoloring and visible artifacts, reflecting their slow convergence. By contrast, ASPOT transfers cool tones more clearly and preserves source structure, producing sharper and more natural images. Overall, ASPOT both converges faster and produces better visual quality than the baselines.

\begin{figure}[t]
    \centering
    \vspace{-2pt}
    \includegraphics[width=0.90\linewidth]
    {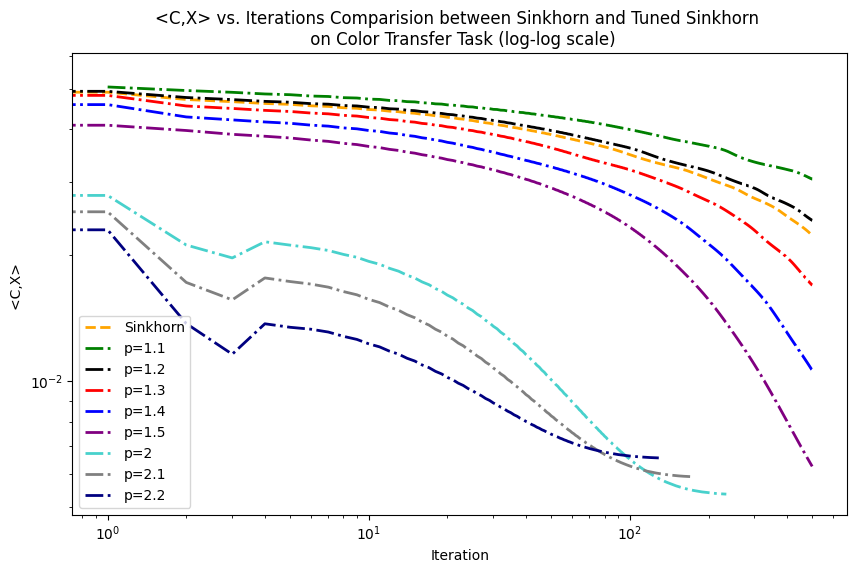}
    \caption{Tuned Sinkhorn Technique Validation}
    \vspace{-4pt}
    \label{fig:tunesink}
\end{figure}

\textbf{Tuned Sinkhorn results} We evaluate tuned Sinkhorn on the color transfer framework described earlier. In this setting, classical Sinkhorn and tuned Sinkhorn technique, use the same tolerance $\texttt{tol}=10^{-2}$, 
and the scaling parameter $\texttt{A\_mult}$ is set to $\|C\|_{\infty}/\texttt{tol}$. 
For the baseline Sinkhorn, $\gamma$ and stopping rule follow the default choice from (Algorithm~\ref{AcceleratedSinkhorn}, Lines~1–8), 
while for tuned Sinkhorn these parameters are selected according to our theoretical prescription in Theorem~\ref{tunedSinkhorn}.  Fig.~\ref{fig:tunesink} reports $\langle C,X\rangle$ against the number of iterations in log-log scale. 
The behavior matches our theoretical insights: as $p$ increases, the convergence of tuned Sinkhorn accelerates noticeably, 
with larger $p$ leading to faster descent. 
Most importantly, by applying the tuned Sinkhorn technique, it consistently outperforms the classical Sinkhorn across all settings. This empirically validates the sharper complexity bound for fine-tuned Sinkhron in Theorem \ref{tunedSinkhorn}.

\subsection{Point Cloud Registration}
\label{sec:pointcloud}
We follow the protocol of \cite{nguyen2024partial, https://doi.org/10.1111/cgf.14614}, registering two 3-D point clouds by applying translations and rotations to align the source cloud with $m$ points to the target cloud with $n$ points. The transported mass is $s=\alpha \cdot \tfrac{\min\{m,n\}}{\max\{m,n\}}$, with $\alpha$ a tunable constant; smaller values improve convergence. In our setup, the blue cloud initially contains 50\% of the red but is misaligned. We set $\alpha=0.4$ for ASPOT and $\alpha=0.7$ for Sinkhorn, with $\gamma=4.4^{-3}$ and annealing rate $0.83$. Different $\alpha$ values are used because Sinkhorn becomes unstable at smaller settings.

\begin{figure}[t]
    \centering
        \vspace{-2pt}
\includegraphics[width=1.05\linewidth]
    {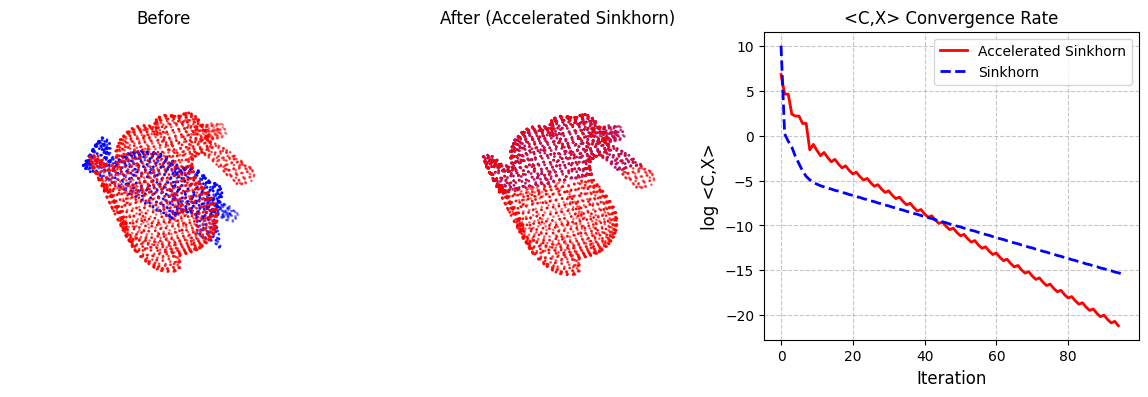}
    \caption{Point Cloud Registration Experiment}
        \vspace{-4pt}
    \label{fig:pointcloud}
\end{figure}

\textbf{Experimental results.}  
ASPOT successfully aligns the two point clouds within 41 registrations, reaching the threshold $10^{-5}$, while Sinkhorn requires 43. Each registration under ASPOT also converges with fewer iterations to compute the POT map. This is illustrated in Fig.~\ref{fig:pointcloud} where the “Before” and “After” visualizations confirm that the transformed blue cloud overlaps tightly with the red one, demonstrating accurate alignment. The right-hand plot further shows that $\langle C,X\rangle$ under ASPOT decreases in an almost straight line on the log scale, reflecting steady reduction of transport cost, whereas Sinkhorn descends more slowly. Taken together, these results highlight two concrete advantages of ASPOT: fewer registrations overall and faster per-registration convergence.

\section{Conclusion}
\label{sec:conclusion}
We developed ASPOT and analyze a tuned varriant technique of Sinkhorn through parameter optimization. Theoretically, we derived better complexity bounds by combining Nesterov-style momentum with Greenkhorn updates and by showing how tuning the regularization parameter improves rates of the classical Sinkhorn. Empirically, color transfer and point cloud experiments confirmed these results: ASPOT converges faster than baselines, while the tuned technique for Sinkhorn consistently improves over the classical version, especially as $p$ increases. We hope these findings can help guide future research on scaling POT to larger and multimarginal settings.

\section{Acknowledgement}  
The authors express their profound gratitude to Mr. Quang Minh Nguyen (MIT, \url{nmquang@mit.edu}) and Mr. Hoang Huy Nguyen (Georgia Tech, \url{hnguyen455@gatech.edu}) for their invaluable input in this project. This research is supported by VNUHCM-University of Information Technology’s Scientific Research Support Fund.

\bibliographystyle{IEEEbib}
\bibliography{strings,reference}

@inproceedings{robust_uot,
title={On Robust Optimal Transport: Computational Complexity and Barycenter Computation},
author={Khang Le and Huy Nguyen and Quang Minh Nguyen and Tung Pham and Hung Bui and Nhat Ho},
booktitle={Advances in Neural Information Processing Systems},
editor={A. Beygelzimer and Y. Dauphin and P. Liang and J. Wortman Vaughan},
year={2021},
url={https://openreview.net/forum?id=xRLT28nnlFV}
}

@article{raghvendra2024new,
  title={A New Robust Partial $ p $-Wasserstein-Based Metric for Comparing Distributions},
  author={Raghvendra, Sharath and Shirzadian, Pouyan and Zhang, Kaiyi},
  journal={arXiv preprint arXiv:2405.03664},
  year={2024}
}

@inproceedings{nguyen2024partial,
  title={On partial optimal transport: Revising the infeasibility of sinkhorn and efficient gradient methods},
  author={Nguyen, Anh Duc and Nguyen, Tuan Dung and Nguyen, Quang Minh and Nguyen, Hoang H and Nguyen, Lam M and Toh, Kim-Chuan},
  booktitle={Proceedings of the AAAI Conference on Artificial Intelligence},
  volume={38},
  pages={8090--8098},
  year={2024}
}

@article{altschuler2017near,
  title={Near-linear time approximation algorithms for optimal transport via Sinkhorn iteration},
  author={Altschuler, Jason and Niles-Weed, Jonathan and Rigollet, Philippe},
  journal={Advances in neural information processing systems},
  volume={30},
  year={2017}
}

@inproceedings{le2022multimarginal,
  title={On multimarginal partial optimal transport: Equivalent forms and computational complexity},
  author={Le, Khang and Nguyen, Huy and Nguyen, Khai and Pham, Tung and Ho, Nhat},
  booktitle={International Conference on Artificial Intelligence and Statistics},
  pages={4397--4413},
  year={2022},
  organization={PMLR}
}

@article{lin2022efficiency,
  title={On the efficiency of entropic regularized algorithms for optimal transport},
  author={Lin, Tianyi and Ho, Nhat and Jordan, Michael I},
  journal={Journal of Machine Learning Research},
  volume={23},
  number={137},
  pages={1--42},
  year={2022}
}

@inproceedings{dvurechensky2018computational,
  title={Computational optimal transport: Complexity by accelerated gradient descent is better than by Sinkhorn’s algorithm},
  author={Dvurechensky, Pavel and Gasnikov, Alexander and Kroshnin, Alexey},
  booktitle={International conference on machine learning},
  pages={1367--1376},
  year={2018},
  organization={PMLR}
}

@misc{mena2019statisticalboundsentropicoptimal,
      title={Statistical bounds for entropic optimal transport: sample complexity and the central limit theorem}, 
      author={Gonzalo Mena and Jonathan Weed},
      year={2019},
      eprint={1905.11882},
      archivePrefix={arXiv},
      primaryClass={math.ST},
      url={https://arxiv.org/abs/1905.11882}, 
}

@misc{balaji2020robustoptimaltransportapplications,
      title={Robust Optimal Transport with Applications in Generative Modeling and Domain Adaptation}, 
      author={Yogesh Balaji and Rama Chellappa and Soheil Feizi},
      year={2020},
      eprint={2010.05862},
      archivePrefix={arXiv},
      primaryClass={cs.LG},
      url={https://arxiv.org/abs/2010.05862}, 
}

@misc{cuturi2013sinkhorndistanceslightspeedcomputation,
      title={Sinkhorn Distances: Lightspeed Computation of Optimal Transportation Distances}, 
      author={Marco Cuturi},
      year={2013},
      eprint={1306.0895},
      archivePrefix={arXiv},
      primaryClass={stat.ML},
      url={https://arxiv.org/abs/1306.0895}, 
}

@article{Khamis_2024,
   title={Scalable Optimal Transport Methods in Machine Learning: A Contemporary Survey},
   ISSN={1939-3539},
   url={http://dx.doi.org/10.1109/TPAMI.2024.3379571},
   DOI={10.1109/tpami.2024.3379571},
   journal={IEEE Transactions on Pattern Analysis and Machine Intelligence},
   publisher={Institute of Electrical and Electronics Engineers (IEEE)},
   author={Khamis, Abdelwahed and Tsuchida, Russell and Tarek, Mohamed and Rolland, Vivien and Petersson, Lars},
   year={2024},
   pages={1–20} }

@article{Wang2022PartialWA,
  title={Partial Wasserstein Adversarial Network for Non-rigid Point Set Registration},
  author={Zi-Ming Wang and Nan Xue and Ling Lei and Guisong Xia},
  journal={ArXiv},
  year={2022},
  volume={abs/2203.02227},
  url={https://api.semanticscholar.org/CorpusID:247244684}
}

@article{JMLR_UOT,
  author  = {Quang Minh Nguyen and Hoang H. Nguyen and Yi Zhou and Lam M. Nguyen},
  title   = {On Unbalanced Optimal Transport: Gradient Methods, Sparsity and Approximation Error},
  journal = {Journal of Machine Learning Research},
  year    = {2023},
  volume  = {24},
  number  = {384},
  pages   = {1--41},
  url     = {http://jmlr.org/papers/v24/22-1158.html}
}

@InProceedings{pmlr-v84-genevay18a,
  title = 	 {Learning Generative Models with Sinkhorn Divergences},
  author = 	 {Genevay, Aude and Peyre, Gabriel and Cuturi, Marco},
  booktitle = 	 {Proceedings of the Twenty-First International Conference on Artificial Intelligence and Statistics},
  pages = 	 {1608--1617},
  year = 	 {2018},
  editor = 	 {Storkey, Amos and Perez-Cruz, Fernando},
  volume = 	 {84},
  series = 	 {Proceedings of Machine Learning Research},
  month = 	 {09--11 Apr},
  publisher =    {PMLR},
  url = 	 {https://proceedings.mlr.press/v84/genevay18a.html}
}

@misc{xie2023solvingspecialtypeoptimal,
      title={Solving a Special Type of Optimal Transport Problem by a Modified Hungarian Algorithm}, 
      author={Yiling Xie and Yiling Luo and Xiaoming Huo},
      year={2023},
      eprint={2210.16645},
      archivePrefix={arXiv},
      primaryClass={math.OC},
      url={https://arxiv.org/abs/2210.16645}, 
}

@article{doi:10.1137/141000439,
author = {Benamou, Jean-David and Carlier, Guillaume and Cuturi, Marco and Nenna, Luca and Peyr\'{e}, Gabriel},
title = {Iterative Bregman Projections for Regularized Transportation Problems},
journal = {SIAM Journal on Scientific Computing},
volume = {37},
number = {2},
pages = {A1111-A1138},
year = {2015},
doi = {10.1137/141000439},

URL = { 
    
        https://doi.org/10.1137/141000439
    
    

},
eprint = { 
    
        https://doi.org/10.1137/141000439
    
    

}
}

@misc{chen2023plotpromptlearningoptimal,
      title={PLOT: Prompt Learning with Optimal Transport for Vision-Language Models}, 
      author={Guangyi Chen and Weiran Yao and Xiangchen Song and Xinyue Li and Yongming Rao and Kun Zhang},
      year={2023},
      eprint={2210.01253},
      archivePrefix={arXiv},
      primaryClass={cs.CV},
      url={https://arxiv.org/abs/2210.01253}, 
}

@inproceedings{bonneel2023survey,
  title={A survey of optimal transport for computer graphics and computer vision},
  author={Bonneel, Nicolas and Digne, Julie},
  booktitle={Computer Graphics Forum},
  volume={42},
  pages={439--460},
  year={2023},
  organization={Wiley Online Library}
}

@misc{phung2025controloptimaltransportneural,
      title={Control, Optimal Transport and Neural Differential Equations in Supervised Learning}, 
      author={Minh-Nhat Phung and Minh-Binh Tran},
      year={2025},
      eprint={2503.15105},
      archivePrefix={arXiv},
      primaryClass={math.NA},
      url={https://arxiv.org/abs/2503.15105}, 
}

@misc{blondel2018smoothsparseoptimaltransport,
      title={Smooth and Sparse Optimal Transport}, 
      author={Mathieu Blondel and Vivien Seguy and Antoine Rolet},
      year={2018},
      eprint={1710.06276},
      archivePrefix={arXiv},
      primaryClass={stat.ML},
      url={https://arxiv.org/abs/1710.06276}, 
}

@inproceedings{NEURIPS2020_1e6e25d9,
 author = {Chapel, Laetitia and Alaya, Mokhtar Z. and Gasso, Gilles},
 booktitle = {Advances in Neural Information Processing Systems},
 editor = {H. Larochelle and M. Ranzato and R. Hadsell and M.F. Balcan and H. Lin},
 pages = {2903--2913},
 publisher = {Curran Associates, Inc.},
 title = {Partial Optimal Tranport with applications on Positive-Unlabeled Learning},
 url = {https://proceedings.neurips.cc/paper_files/paper/2020/file/1e6e25d952a0d639b676ee20d0519ee2-Paper.pdf},
 volume = {33},
 year = {2020}
}

@article{
kemertas2025efficient,
title={Efficient and Accurate Optimal Transport with Mirror Descent and Conjugate Gradients},
author={Mete Kemertas and Allan Douglas Jepson and Amir-massoud Farahmand},
journal={Transactions on Machine Learning Research},
issn={2835-8856},
year={2025},
url={https://openreview.net/forum?id=FVFqrxeF8e},
note={}
}

@article{https://doi.org/10.1111/cgf.14614,
author = {Qin, Hongxing and Zhang, Yucheng and Liu, Zhentao and Chen, Baoquan},
title = {Rigid Registration of Point Clouds Based on Partial Optimal Transport},
journal = {Computer Graphics Forum},
volume = {41},
number = {6},
pages = {365-378},
doi = {https://doi.org/10.1111/cgf.14614},
url = {https://onlinelibrary.wiley.com/doi/abs/10.1111/cgf.14614},
eprint = {https://onlinelibrary.wiley.com/doi/pdf/10.1111/cgf.14614},
year = {2022}
}

@inproceedings{DBLP:conf/eccv/ZhengHTX24,
  author={Mengyu Zheng and Zhiwei Hao and Yehui Tang and Chang Xu},
  title={Visual Prompting via Partial Optimal Transport},
  year={2024},
  cdate={1704067200000},
  pages={1-18},
  url={https://doi.org/10.1007/978-3-031-72761-0_1},
  booktitle={ECCV (35)},
}

@inproceedings{
zhang2024pot,
title={P\${\textasciicircum}2\${OT}: Progressive Partial Optimal Transport for Deep Imbalanced Clustering},
author={Chuyu Zhang and Hui Ren and Xuming He},
booktitle={The Twelfth International Conference on Learning Representations},
year={2024},
url={https://openreview.net/forum?id=hD3sGVqPsr}
}

@article{DBLP:journals/corr/abs-2404-03446,
  publtype={informal},
  author={Chuyu Zhang and Hui Ren and Xuming He},
  title={SPOT: Semantic-Regularized Progressive Partial Optimal Transport for Imbalanced Clustering},
  year={2024},
  cdate={1704067200000},
  journal={CoRR},
  volume={abs/2404.03446},
  url={https://doi.org/10.48550/arXiv.2404.03446}
}

@phdthesis{papadakis:tel-01246096,
  TITLE = {{Optimal Transport for Image Processing}},
  AUTHOR = {Papadakis, Nicolas},
  URL = {https://hal.science/tel-01246096},
  SCHOOL = {{Universit{\'e} de Bordeaux ; Habilitation thesis}},
  YEAR = {2015},
  MONTH = Dec,
  TYPE = {Habilitation {\`a} diriger des recherches},
  HAL_ID = {tel-01246096},
  HAL_VERSION = {v8},
}

@ARTICLE{7974883,
  author={Kolouri, Soheil and Park, Se Rim and Thorpe, Matthew and Slepcev, Dejan and Rohde, Gustavo K.},
  journal={IEEE Signal Processing Magazine}, 
  title={Optimal Mass Transport: Signal processing and machine-learning applications}, 
  year={2017},
  volume={34},
  number={4},
  pages={43-59},
  doi={10.1109/MSP.2017.2695801}}

@book{villani2008optimal,
  title={Optimal transport: old and new},
  author={Villani, C{\'e}dric and others},
  volume={338},
  year={2008},
  publisher={Springer}
}

@Article{Kantorovich2006,
author={Kantorovich, L. V.},
title={On the Translocation of Masses},
journal={Journal of Mathematical Sciences},
year={2006},
month={Mar},
day={01},
volume={133},
number={4},
pages={1381-1382},
issn={1573-8795},
doi={10.1007/s10958-006-0049-2},
url={https://doi.org/10.1007/s10958-006-0049-2}
}

@book{monge1781mémoire,
  title={M{\'e}moire sur la th{\'e}orie des d{\'e}blais et des remblais},
  author={Monge, G.},
  url={https://books.google.com/books?id=IG7CGwAACAAJ},
  year={1781},
  publisher={Imprimerie royale}
}

@misc{chizat2017scalingalgorithmsunbalancedtransport,
      title={Scaling Algorithms for Unbalanced Transport Problems}, 
      author={Lenaic Chizat and Gabriel Peyré and Bernhard Schmitzer and François-Xavier Vialard},
      year={2017},
      eprint={1607.05816},
      archivePrefix={arXiv},
      primaryClass={math.OC},
      url={https://arxiv.org/abs/1607.05816}, 
}

@misc{deplaen2023unbalancedoptimaltransportunified,
      title={Unbalanced Optimal Transport: A Unified Framework for Object Detection}, 
      author={Henri De Plaen and Pierre-François De Plaen and Johan A. K. Suykens and Marc Proesmans and Tinne Tuytelaars and Luc Van Gool},
      year={2023},
      eprint={2307.02402},
      archivePrefix={arXiv},
      primaryClass={cs.CV},
      url={https://arxiv.org/abs/2307.02402}, 
}

@misc{li2025optimaltransportbasedgenerativemodels,
      title={Optimal Transport-Based Generative Models for Bayesian Posterior Sampling}, 
      author={Ke Li and Wei Han and Yuexi Wang and Yun Yang},
      year={2025},
      eprint={2504.08214},
      archivePrefix={arXiv},
      primaryClass={stat.CO},
      url={https://arxiv.org/abs/2504.08214}, 
}

@misc{courty2016optimaltransportdomainadaptation,
      title={Optimal Transport for Domain Adaptation}, 
      author={Nicolas Courty and Rémi Flamary and Devis Tuia and Alain Rakotomamonjy},
      year={2016},
      eprint={1507.00504},
      archivePrefix={arXiv},
      primaryClass={cs.LG},
      url={https://arxiv.org/abs/1507.00504}, 
}

@article{Pham2024,
title = {Application of Unbalanced Optimal Transport in Healthcare},
journal = {International Journal of Advanced Computer Science and Applications},
doi = {10.14569/IJACSA.2024.0151110},
url = {http://dx.doi.org/10.14569/IJACSA.2024.0151110},
year = {2024},
publisher = {The Science and Information Organization},
volume = {15},
number = {11},
author = {Qui Phu Pham and Nghia Thu Truong and Hoang-Hiep Nguyen-Mau and Cuong Nguyen and Mai Ngoc Tran and Dung Luong}
}

\section{Additional Literature Review}
Optimal Transport (OT), introduced by Monge~\cite{monge1781mémoire} and formalized by Kantorovich~\cite{Kantorovich2006}, seeks the cost-minimizing way to reallocate mass from a source distribution to a target according to a given ground cost.  Building on this foundation, Villani’s influential monograph~\cite{villani2008optimal} established OT as a cornerstone in modern mathematics, bridging theory and applications. More recently, the field experienced a surge thanks to computational breakthroughs where a line of scalable algorithms has been developed, including interior-point and combinatorial solvers~\cite{doi:10.1137/141000439}, the entropically regularized Sinkhorn algorithm~\cite{cuturi2013sinkhorndistanceslightspeedcomputation}, accelerated gradient-based methods~\cite{dvurechensky2018computational}, efficiency improvements for large-scale problems~\cite{lin2022efficiency}, and specialized schemes for structured OT settings~\cite{xie2023solvingspecialtypeoptimal}. In parallel, entropic regularization~\cite{balaji2020robustoptimaltransportapplications,cuturi2013sinkhorndistanceslightspeedcomputation,lin2022efficiency,pmlr-v84-genevay18a,mena2019statisticalboundsentropicoptimal} has proven particularly powerful: beyond improving statistical properties and mitigating the curse of dimensionality, it renders OT differentiable with respect to the input distributions, making it naturally compatible with modern deep learning frameworks~\cite{bonneel2023survey,chen2023plotpromptlearningoptimal}. As a result, OT has been adopted across a wide spectrum of applications, from economics and control to machine learning, Bayesian statistics, and signal processing~\cite{li2025optimaltransportbasedgenerativemodels,dvurechensky2018computational,pmlr-v84-genevay18a,Khamis_2024,phung2025controloptimaltransportneural,Wang2022PartialWA,papadakis:tel-01246096,7974883}. 

Despite these advances, classical OT suffers from two fundamental limitations. First, it requires the two distributions to have equal total mass, which can be prohibitive in many practical scenarios where this assumption does not hold. Second, OT is sensitive to outliers and sampling discrepancies, issues that often arise when computing Wasserstein distances from real data~\cite{raghvendra2024new}. To overcome these barriers, researchers have proposed several generalizations. Unbalanced Optimal Transport (UOT) relaxes the mass constraint by introducing divergence penalties, with recent contributions offering both theoretical insights and unified frameworks~\cite{JMLR_UOT,chizat2017scalingalgorithmsunbalancedtransport,deplaen2023unbalancedoptimaltransportunified, Pham2024}. Partial Optimal Transport (POT), on the other hand, imposes an explicit cap on the transported mass, making the formulation more robust to noise and better suited for tasks with mismatched supports \cite{DBLP:journals/corr/abs-2404-03446,DBLP:conf/eccv/ZhengHTX24,zhang2024pot,le2022multimarginal,nguyen2024partial,NEURIPS2020_1e6e25d9}.

The trajectory of POT research highlights an evolving landscape. Early approaches, such as the reformulation strategy of Chapel and colleagues, cast POT as a variant of OT by augmenting marginals with dummy mass \cite{NEURIPS2020_1e6e25d9}. This perspective was refined by ~\cite{le2022multimarginal}, who provided the first theoretical complexity bound of $\tilde{O}(n^2/\varepsilon^2)$. However, ~\cite{nguyen2024partial} later demonstrated that the rounding procedure in fact yields infeasible solutions, casting doubt on the applicability of Sinkhorn-based methods for POT. \cite{nguyen2024partial} also proposed an accelerated first-order methods namely APDAGD provide provable rates but incur higher complexity of $\tilde{O}(n^{2.5}/\varepsilon)$, making them less practical for large-scale problems. 
In parallel, the growing list of applications has underscored the importance of POT as a modeling tool. Its robustness to outliers and partial matching has been leveraged in computer vision and graphics~\cite{DBLP:journals/corr/abs-2404-03446,DBLP:conf/eccv/ZhengHTX24,zhang2024pot}, and in robust machine learning tasks such as domain adaptation, point cloud registration, and color transfer~\cite{nguyen2024partial,NEURIPS2020_1e6e25d9}. These diverse applications demonstrate that while classical OT and UOT each provide valuable frameworks, POT uniquely addresses the challenges of partial alignment, establishing itself as a crucial frontier where theory and practice continue to meet.

\section{APPENDIX}
\label{sec:related_works}
\label{sec:appendix}
\subsection{Reformulating POT as Extended OT}
It has been shown in \cite{nguyen2024partial, NEURIPS2020_1e6e25d9} that the POT problem can be reformulated as a standard OT problem by introducing slack variables and dummy nodes. For completeness, we briefly restate the construction here.

\textbf{Linear Programming Form of  POT.} We first introduce slack variables $p,q \in \mathbb{R}^n_{+}$ to capture the unmatched mass on the source and target sides. The constraints of POT can then be written explicitly as
\begin{align}
    X\mathbf{1}_n + p = r, 
    \quad 
    X^\top \mathbf{1}_n + q = c, 
    \quad 
    \mathbf{1}_n^\top X \mathbf{1}_n = s,
\end{align}
where $X \in \mathbb{R}^{n\times n}_{+}$ is the transport plan, $r,c \in \mathbb{R}^n_{+}$ are the marginals, and $s$ is the transport budget.  

This system can be compactly encoded in canonical linear programming (LP) form:
\begin{align}
    \min_{x \geq 0}\;\langle d, x\rangle \quad \text{s.t. } Ax=b,
\end{align}
where the combined decision variable $x \in \mathbb{R}_{+}^{n^2+2n}$ stacks the transport plan and slack variables,
\begin{align}
    x=\begin{pmatrix}
    \operatorname{vec}(X)\\ p \\ q
    \end{pmatrix},
    \quad 
    d=\begin{pmatrix}
    \operatorname{vec}(C)\\ \textbf{0}_{2n}
    \end{pmatrix},
\end{align}
with $\operatorname{vec}(X)$ denoting the column-stacking of $X$ into a vector and $\textbf{0}_{n}$ is the $n$-vector of zeros.
The linear operator $A$ and right-hand side $b$ are defined as
\begin{align}
    A=\begin{pmatrix}
    A' & I_{2n}\\
    \mathbf{1}_{n^2}^\top & \textbf{0}_{2n}
    \end{pmatrix},
    \quad
    b=\begin{pmatrix}
    r \\ c \\ s
    \end{pmatrix},
\end{align}
where $A' \in \mathbb{R}^{2n \times n^2}$ is the standard node–edge incidence matrix used in OT problems \cite{dvurechensky2018computational}. The last row ensures that the total transported mass equals $s$.

\textbf{Reduction to Extended OT}
To leverage efficient OT solvers, particularly those based on entropic regularization, we further reformulate POT as a balanced OT problem by introducing dummy points. Specifically, we extend the cost matrix to dimension $(n+1)\times(n+1)$:
\begin{align}
\label{C_tilde}
\widetilde{C}=\begin{pmatrix}
C & \textbf{0}_n\\
\textbf{0}_n^\top & A
\end{pmatrix},
\end{align}
where $A > \max_{i,j} C_{ij}$ guarantees that sending mass to the dummy node is strictly more expensive than any feasible transport \cite{NEURIPS2020_1e6e25d9}.  

The marginals are augmented accordingly:
\begin{align}
\label{eq:r,c}
\tilde{r}=\begin{pmatrix} r \\ \|c\|_1 - s \end{pmatrix},
\quad
\tilde{c}=\begin{pmatrix} c \\ \|r\|_1 - s \end{pmatrix},
\end{align}
so that $\|\tilde{r}\|_1 = \|\tilde{c}\|_1$.

Any feasible transport plan $\widetilde{X}\in\mathbb{R}_{+}^{(n+1)\times(n+1)}$ for this OT problem can be written as
\begin{align}
\label{X_tilde}
\widetilde{X}=\begin{pmatrix}
\bar{X} & \tilde{p}\\
\tilde{q}^\top & \widetilde{X}_{n+1,n+1}
\end{pmatrix},
\end{align}
where $\bar{X}\in\mathbb{R}_{+}^{n\times n}$ is the transport restricted to the original support, and $(\tilde{p},\tilde{q})$ correspond to flows involving the dummy nodes. By construction, $\bar{X}$ is a feasible solution to the original POT problem \cite{NEURIPS2020_1e6e25d9}.

\subsection{Proof of Lemma 1}
\begin{proof}
Recall that if $f$ has a Lipschitz-continuous gradient with constant $L>0$, then
\begin{align}
    f(y)-f(x) -\langle \nabla f(x), y-x\rangle \leq \tfrac{L}{2}\|x-y\|^2. 
\end{align}
Applying this property and the relations from Algorithm \ref{AcceleratedSinkhorn}, we obtain
\begin{align}
\label{AccelProof}
\varphi(\grave{z}^t) \leq \varphi(\bar{z}^t) 
+ \theta_t (\tilde{z}^{t+1}-\tilde{z}^t)^\top \nabla\varphi(\bar{z}^t)
+ \tfrac{L}{2}\theta_t^2 \|\tilde{z}^{t+1}-\tilde{z}^t\|^2.
\end{align}

On the other hand, we note
\begin{align}
\label{simpleCalculations1}
\varphi(\bar{z}^t) = (1-\theta_t)\varphi(\bar{z}^t)+\theta_t\varphi(\bar{z}^t),
\end{align}
and
\begin{align}
\label{simpleCalculations2}
(\tilde{z}^{t+1}-\tilde{z}^t)^\top \nabla\varphi(\bar{z}^t)
&= -(\tilde{z}^t-\bar{z}^t)^\top \nabla\varphi(\bar{z}^t)
\notag \\&+ (\tilde{z}^{t+1}-\bar{z}^t)^\top \nabla\varphi(\bar{z}^t).
\end{align}
Plugging \eqref{simpleCalculations1} and \eqref{simpleCalculations2} into \eqref{AccelProof} gives
\begin{align}
\label{2 term}
\varphi(\grave{z}^t) &\leq 
\theta_t \underbrace{\Big( \varphi(\bar{z}^t)+(\tilde{z}^{t+1}-\bar{z}^t)^\top\nabla\varphi(\bar{z}^t) \Big)}_{\text{Term 1}}  \notag
\\&+ \underbrace{(1-\theta_t)\Big(\varphi(\bar{z}^t)-(\tilde{z}^t-\bar{z}^t)^\top\nabla\varphi(\bar{z}^t)\Big)}_{\text{Term 2  }}.
\end{align}

From Algorithm \ref{AcceleratedSinkhorn}, for every $z \in \mathbb{R}^{2n+1}$,
\begin{align}
\label{equals0}
(z-\tilde{z}^{t+1})^\top \big[\nabla\varphi(\bar{z}^t)+L\theta_t(\tilde{z}^{t+1}-\tilde{z}^t)\big]=0.
\end{align}
Also note the convexity inequality
\begin{align}
\label{convexity-z}
\big(z^{\star}-\bar{z}^{\,t}\big)^\top \nabla\varphi(\bar{z}^{\,t})
\;\le\; \varphi(z^{\star})-\varphi(\bar{z}^{\,t}).
\end{align}
In \eqref{equals0}, substitute $z=z^{\star}$ to obtain
\begin{align}
\label{general-z}
&(\tilde{z}^{\,t+1}-\bar{z}^{\,t})^\top \nabla\varphi(\bar{z}^{\,t})
= (z^{\star}-\bar{z}^{\,t})^\top \nabla\varphi(\bar{z}^{\,t}) \notag
\\&+\frac{L}{2}\theta_t\!\left(\|z^{\star}-\tilde{z}^{\,t}\|^2-\|z^{\star}-\tilde{z}^{\,t+1}\|^2-\|\tilde{z}^{\,t+1}-\tilde{z}^{\,t}\|^2\right).
\end{align}
Combining \eqref{convexity-z} and \eqref{general-z} yields the bound for Term 1 in \eqref{2 term}:
\begin{align}
\label{term1-z}
\text{Term 1}
\;\le\; \varphi(z^{\star})
+\frac{L}{2}\theta_t\!\left(\|z^{\star}-\tilde{z}^{\,t}\|^2-\|z^{\star}-\tilde{z}^{\,t+1}\|^2\right).
\end{align}

Next, observe the interpolation identity implied by the algorithmic relations,
\begin{align*}
-\theta_t(\tilde{z}^{\,t}-\bar{z}^{\,t})
&= \theta_t \bar{z}^{\,t}+(1-\theta_t)\check{z}^{\,t} - \bar{z}^{\,t} \notag  \\&= (1-\theta_t)(\check{z}^{\,t}-\bar{z}^{\,t}),
\end{align*}
which transforms Term 2 in \eqref{2 term} as
\begin{align}
\label{term2-z}
\text{Term 2}
\;\le\; (1-\theta_t)\Big(\varphi(\bar{z}^{\,t})+(\check{z}^{\,t}-\bar{z}^{\,t})^\top\nabla\varphi(\bar{z}^{\,t})\Big).
\end{align}
Putting \eqref{term1-z} and \eqref{term2-z} into \eqref{2 term}, we obtain
\begin{align}
\notag
\varphi(\grave{z}^{\,t})
&\le (1-\theta_t)\varphi(\check{z}^{\,t})+\theta_t\varphi(z^{\star}) \notag
\\&+\frac{L}{2}\theta_t^2\!\left(\|z^{\star}-\tilde{z}^{\,t}\|^2-\|z^{\star}-\tilde{z}^{\,t+1}\|^2\right).
\end{align}

Finally, by the descent property of the Greenkhorn updates in Algorithm~\ref{AcceleratedSinkhorn},
\[
\varphi(\grave{z}^{\,t}) \;\ge\; \varphi(\hat{z}^{\,t})
\;\ge\; \varphi(z^{\,t})
\;\ge\; \varphi(\check{z}^{\,t+1}),
\]
and therefore
\begin{align*}
\varphi(\check{z}^{\,t+1})-\varphi(z^{\star})
&\le (1-\theta_t)\big(\varphi(\check{z}^{\,t})-\varphi(z^{\star})\big) \notag
\\&+\frac{L}{2}\theta_t^2\!\left(\|z^{\star}-\tilde{z}^{\,t}\|^2-\|z^{\star}-\tilde{z}^{\,t+1}\|^2\right),
\end{align*}
which is exactly
\[
\delta_{t+1} \;\le\; (1-\theta_t)\,\delta_t
+ \tfrac{L}{2}\theta_t^2\Big(\|z^{\star}-\tilde{z}^{\,t}\|^2-\|z^{\star}-\tilde{z}^{\,t+1}\|^2\Big).
\]
\end{proof}
\subsection{Proof of Lemma 2}
\begin{proof}
From the definition of $\theta_t$ we know that
\[
\frac{\theta_{t+1}}{\theta_t} = \sqrt{1-\theta_{t+1}}.
\]
Therefore, applying Lemma \ref{lemma:1}, we obtain
\begin{align}
   \left(\frac{1-\theta_{t+1}}{\theta_{t+1}^2}\right)\delta_{t+1}
  & -\left(\frac{1-\theta_t}{\theta_t^2}\right)\delta_t \notag
  \\& \;\leq\; \frac{L}{2}\Big(\|z^{\star}-\tilde{z}^{\,t}\|^2 - \|z^{\star}-\tilde{z}^{\,t+1}\|^2\Big).
\end{align}

Unfolding this inequality recursively gives
\begin{align}
\notag
\left(\frac{1-\theta_t}{\theta_t^2}\right)\delta_t
+ \frac{L}{2}\|z^{\star}-\tilde{z}^{\,t}\|^2
&\;\leq\; \left(\frac{1-\theta_0}{\theta_0^2}\right)\delta_0
+ \frac{L}{2}\|z^{\star}-\tilde{z}^{\,0}\|^2 \\
&=\; \frac{L}{2}\|z^{\star}\|^2,
\end{align}
since $\tilde{z}^{\,0}=0$ and $\delta_0=0$.

Thus,
\begin{align}
\delta_t \;\leq\; \tfrac{L}{2}\theta_{t-1}^2 \|z^{\star}\|^2
\;\leq\; L\,\theta_{t-1}^2\,nR^2,
\end{align}
where we used $\|z^{\star}\|\leq \sqrt{2n}\,R$.

Finally, it is known from \cite{lin2022efficiency} that $0<\theta_t\leq \tfrac{2}{t+2}$, which leads to the desired bound
\[
\delta_t \;\leq\; \frac{4nLR^2}{(t+1)^2}.
\]
\end{proof}
\subsection{Proof of Theorem 3}
In this proof, we denote the row and column sums of a matrix $A$ by  $r(A) = A \mathbf{1}_n$ and $c(A) = A^\top \mathbf{1}_n$.
\begin{proof}
Define
\[
S = \varphi(z^t) - \varphi(\check{z}^{\,t+1}).
\]
Suppose the update is on the $u_i$-coordinate (the case for $v_j$ is similar). Then
\begin{align}
\notag
S &= \underbrace{\sum_{i,j=1}^{n} e^{\left(-\tfrac{C_{ij}}{\gamma} + u_i^t + v_j^t + w^t\right)} + \sum_{i=1}^n e^{u_i^t}}_{\text{Term I}}
 - \sum_{i=1}^n u_i^t r_i \\
&\quad -\Bigg(
\underbrace{\sum_{i,j=1}^{n} e^{\left(-\tfrac{C_{ij}}{\gamma} + \check{u}_i^{t+1} + \check{v}_j^{t+1} + \check{w}^{t+1}\right)} + \sum_{i=1}^n e^{\check{u}_i^{t+1}}}_{\text{Term II}}
 - \sum_{i=1}^n \check{u}_i^{t+1} r_i\Bigg).
\end{align}

From Algorithm \ref{AcceleratedSinkhorn}, using the relation between $\check{u}^{t+1}$, $\tilde{p}^t$, and $u^t$, we obtain
\begin{align}
\label{sum}
\sum_{i=1}^n \check{u}_i^{t+1} r_i - \sum_{i=1}^n u_i^t r_i
= \sum_{i=1}^n r_i \log\!\left(\frac{r_i}{r_i(B(z^t))+\tilde{p}_i^t}\right).
\end{align}
Meanwhile,
\begin{align}
\label{sum row}
\text{Term I} = \sum_{i=1}^n \big(r_i(B(z^t))+\tilde{p}_i^t\big).
\end{align}
Since $\frac{\partial \varphi}{\partial u_i}=0$, it holds that
\[
\sum_{j=1}^n e^{\left(-\tfrac{C_{ij}}{\gamma}+u_i+v_j+w\right)} + e^{u_i} = r_i,
\]
so that Term II $=\sum_{i=1}^n r_i$. Now from \ref{sum}, \ref{sum row}, we arrive at,
\begin{align}
\label{update u}
S = \sum_{i=1}^n \rho\!\left(r_i,\, r_i(B(z^t))+\tilde{p}_i^t\right).
\end{align}

If the update is instead on $w$, we obtain
\begin{align}
S &= \underbrace{\sum_{i,j=1}^n e^{\left(-\tfrac{C_{ij}}{\gamma}+u_i^t+v_j^t+w^t\right)}}_{\text{Term I}} - w^t s \notag
\\& -\Bigg(\underbrace{\sum_{i,j=1}^n e^{\left(-\tfrac{C_{ij}}{\gamma}+\check{u}_i^{t+1}+\check{v}_j^{t+1}+\check{w}^{t+1}\right)}}_{\text{Term II}} - w^{t+1}s\Bigg).
\end{align}
From Algorithm \ref{AcceleratedSinkhorn}, the relation between $\check{w}^{t+1}$ and $w^t$ gives
\[
w^{t+1}s - w^t s = s\log\!\left(\frac{s}{\|B(z^t)\|_1}\right).
\]
Also,
\begin{align}
\notag
\text{Term I} - \text{Term II} 
&= \sum_{i,j=1}^n e^{\left(-\tfrac{C_{ij}}{\gamma}+u_i^t+v_j^t+w^t\right)}
\left(1-\frac{s}{\|B(z^t)\|_1}\right) \\
&= \|B(z^t)\|_1 - s. \notag
\end{align}
Thus,
\begin{align}
    \label{update w}
    S = \rho\!\left(s,\, \|B(z^t)\|_1\right).
\end{align}
Combining the cases in \ref{update u} and \ref{update w}, we obtain
\begin{align}
\notag
S &\geq \tfrac{1}{3}\sum_{i=1}^n \rho\!\left(r_i,\, r_i(B(z^t))+\tilde{p}_i^t\right)
+ \tfrac{1}{3}\sum_{j=1}^n \rho\!\left(c_j,\, c_j(B(z^t))+\tilde{q}_j^t\right) \\
&\quad+ \tfrac{1}{3}\rho\!\left(s,\, \|B(z^t)\|_1\right). \notag
\end{align}
By Lemma 6 in \cite{altschuler2017near} and the Cauchy–Schwarz inequality,
\begin{align}
\notag
S &\geq \tfrac{1}{21}\!\Big(
   \big|\mathbf{1}_n^\top B^t \mathbf{1}_n - s\big|^2 \\
   &\quad+ \|B^t \mathbf{1}_n + \tilde{p} - r\|_1^2
   + \|(B^t)^\top \mathbf{1}_n + \tilde{q} - c\|_1^2
\Big) \notag \\ 
&\geq \tfrac{1}{63}\big(E(t)\big)^2. \notag
\end{align}

Since $\varphi(\check{z}^t)\geq \varphi(\check{z}^{\,t+1})$, it follows that
\[
\varphi(\check{z}^t)-\varphi(\check{z}^{\,t+1})
\;\geq\; \tfrac{1}{63}\sum_{i=j}^t E_i^2
\quad\text{for any } j\in\{1,\dots,t\}.
\]
On the other hand, 
\[
\varphi(\check{z}^j)-\varphi(\check{z}^{\,t+1}) \leq \delta_j.
\]
Combining this with Lemma \ref{lemma:2} yields
\[
\sum_{i=j}^t E_i^2 \;\leq\; \frac{252\,nLR^2}{(j+1)^2}.
\]

Now, since $E_t \geq \varepsilon^{\prime}$,
\[
\frac{252\,nLR^2}{(j+1)^2 (t-j+1)} \;\geq\; (\varepsilon^{\prime})^2
\quad \text{for any } j \in \{1,\dots,t\}.
\]
Without loss of generality, assume $t$ is even and set $j = t/2$. Then
\[
t \;\leq\; 1 + \left(\frac{12 \sqrt{14nL}\,R}{\varepsilon^{\prime}}\right)^{2/3}.
\]
This completes the proof.
\end{proof}
\subsection{Proof of Theorem 1}
\begin{proof}
In this proof, we denote by $\bar{X}$ the rounded plan produced by Algorithm~\ref{AcceleratedSinkhorn}, and by $X^{\star}$ the optimal plan of the original POT problem.  
We also write $\tilde{X}$ for the entropic OT solution with extended cost $\tilde{C}$, and $\tilde{X}^{\star}$ for its optimum.

From Theorem~5.4 of \cite{le2022multimarginal},  
\[
\langle \tilde{C},\tilde{X}\rangle - \langle \tilde{C},\tilde{X}^{\star}\rangle \leq \varepsilon.
\]
Moreover, \cite{NEURIPS2020_1e6e25d9} shows that $\langle \tilde{C},\tilde{X}^{\star}\rangle = \langle C,X^{\star}\rangle$.  
Since simple rounding yields $\langle C,\bar{X}\rangle \leq \langle \tilde{C},\tilde{X}\rangle$, we obtain
\[
\langle C,\bar{X}\rangle - \langle C,X^{\star}\rangle \leq \varepsilon,
\]
so the approximation guarantee holds.

Next, by Theorem~\ref{bound iterations for Accel}, the iteration count satisfies
\[
t \leq 1+\Bigg(\frac{12\sqrt{14nL}\,R}{\varepsilon'}\Bigg)^{2/3},
\]
where $\varepsilon'=\varepsilon/(8\|C\|_{\infty})$, $R$ is defined in \ref{eq:R}, and $\gamma=\varepsilon/(4\log n)$ is the regularization parameter.  
Substituting these definitions, we find
\begin{align*}
t &\leq 1 + \Bigg( 
   \frac{96\sqrt{14nL}\,\|C\|_{\infty}}{\varepsilon}
   \Bigg[
     \frac{4\log(n)\,\|C\|_{\infty}\max\{\|r\|_1,\|c\|_1\}}
          {\varepsilon\big(\max\{\|r\|_1,\|c\|_1\}-s\big)} \\[-2pt]
   &\qquad\qquad
     - \log\!\big(\min_{i,j}(r_i,c_j)\big)
   \Bigg]
   \Bigg)^{2/3}.
\end{align*}

Finally, inserting the Lipschitz constant 
\[
L=\frac{12\log(n)(\|r\|_1+\|c\|_1-s)}{\varepsilon},
\]
yields
\[
t=\mathcal{O}\!\left(\frac{n^{1/3}\,\|C\|_{\infty}^{4/3}(\log n)^{1/3}}{\varepsilon^{5/3}}\right).
\]

Since each iteration of Algorithm~\ref{AcceleratedSinkhorn} costs $\mathcal{O}(n^2)$ arithmetic operations, the overall complexity is
\[
\mathcal{O}\!\left(\frac{n^{7/3}\,\|C\|_{\infty}^{4/3}(\log n)^{4/3}}{\varepsilon^{5/3}}\right).
\]
\end{proof}

\subsection{Proof of Theorem 2}
\begin{proof}
Let $\bar{X}$ be the rounded solution of POT and let $\bar{X}^{\star}$ be an optimal plan.  
We embed POT into the extended $(n+1)\times(n+1)$ OT formulation \cite{NEURIPS2020_1e6e25d9}, whose entropic solution we denote by $\tilde{X}$ and optimum by $\tilde{X}^{\star}$.  
The top-left $n\times n$ blocks of these matrices are written $\tilde{X}_{\oplus}$ and $\tilde{X}^\star_{\oplus}$.

The cost difference can then be split as
\[
\langle \bar{X}-\bar{X}^{\star},C\rangle
= \langle \bar{X}-\tilde{X}_{\oplus},C\rangle
+ \langle \tilde{X}_{\oplus}-\tilde{X}^\star_{\oplus},C\rangle
+ \langle \tilde{X}^\star_{\oplus}-\bar{X}^{\star},C\rangle.
\]
The first term comes from rounding and, by Theorem 6 of \cite{nguyen2024partial}, contributes at most $23\varepsilon'$.  
The last term reflects the difference between the block $\tilde{X}^\star_X$ and the true POT optimum, which Proposition 1 of \cite{NEURIPS2020_1e6e25d9} bounds by $\tfrac{1}{2}\varepsilon'$.  
The remaining term measures the gap between the entropic solution $\tilde{X}_{\oplus}$ and the block of the OT optimum $\tilde{X}^\star_{\oplus}$.  
From \cite{kemertas2025efficient} we know that
\[
\langle \tilde{X}-\tilde{X}^{\star},\tilde{C}\rangle 
   \leq H_{\min}(\tilde r,\tilde c)\,\gamma^p,
\]
where $\tilde{r}, \tilde{c}$ are defined in \ref{eq:r,c} and $\tilde{C}$ is defined in \ref{C_tilde} . And based on its matrix structure in \ref{C_tilde}, \ref{X_tilde} , we immediately have
\begin{align}
\langle \tilde{X}-\tilde{X}^\star,\tilde{C}\rangle
= \langle \tilde{X}_{\oplus}-\tilde{X}^\star_{\oplus},C\rangle
+ \big(\tilde{X}_{n+1,n+1}-\tilde{X}^\star_{n+1,n+1}\big)A. \notag
\end{align}
Since $\tilde{X}^\star_{n+1,n+1}=0$ by \cite{kemertas2025efficient} and specifically \cite{nguyen2024partial} proved that
\[
\tilde{X}_{n+1,n+1} \;\geq\; 
\exp\!\left(-\tfrac{A}{\gamma}-\|\mathbf{u}\|_{\infty}-\|\mathbf{v}\|_{\infty}\right),
\]
together with the dual bound in \cite{nguyen2024partial}
\[
\|\mathbf{u}\|_{\infty},\|\mathbf{v}\|_{\infty} \leq 
 \tfrac{A}{\gamma}+\log n-2\log\!\big(\min_{i,j}\{r_i,c_j\}\big).
\]
This implies that $\tilde{X}_{n+1,n+1} \geq K \exp(-3A/\gamma)$ for some universal constant $K>0$, and therefore the matrix error satisfies
\begin{align}
    \langle \tilde{X}_{\oplus}-\tilde{X}^\star_{\oplus},C\rangle
   \leq H_{\min}(\tilde r,\tilde c)\,\gamma^p - K A \exp(-3A/\gamma). \notag
\end{align}

Putting everything together, we obtain
\[
\langle \bar{X}-\bar{X}^{\star},C\rangle
   \leq \tfrac{47}{2}\varepsilon'
      + H_{\min}(\tilde r,\tilde c)\,\gamma^p
      - K A \exp(-3A/\gamma).
\]
Following the similar argument in \cite{kemertas2025efficient}, we now set 
$\varepsilon' = H_{\min}(r,c)\gamma^p$, 
$A = \tilde{O}(\|C\|_{\infty}/\varepsilon)$, 
and 
$\gamma = (2\varepsilon / (49H_{\min}(r,c)))^{1/p}$.  
With this choice, the total error is at most $\varepsilon+\tilde{O}(\varepsilon^2)$.

Finally, each Sinkhorn iteration costs $O(n^2)$, and the number of iterations required is $\mathcal{O}(A^2/(\gamma\varepsilon'))$.  
Thus the overall runtime is $\mathcal{O}\!\left(\tfrac{n^2 A^2}{\gamma \varepsilon'}\right)$,
which, after substitution, gives the claimed complexity $\mathcal{O}\!\left(
   \frac{n^2 H_{\min}(r,c)^{1/p}\,\|C\|_{\infty}^2}{\varepsilon^{3+1/p}}
\right).$ And when $p \rightarrow \infty$, we conclude our complexity is  $\mathcal{O}\left(n^2H_{\text{min}}(r,c)^{1/p}\|C\|^2_{\infty}/\varepsilon^{3+o(1)}\right).$
\end{proof}
\subsection{Further experimental discussion}
\label{app:further-exp}
\textbf{ASPOT Synthetic Data Experiment.} We provide additional experiments to evaluate the scalability of ASPOT in terms of wall-clock runtime. Specifically, we generate two random discrete probability distributions: the source distribution $x \in \mathbb{R}^n$ and the target distribution $y \in \mathbb{R}^n$, where both are non-negative vectors normalized such that $\|x\|_{1} = 5$ and $\|y\|_{1} = 3$. The cost matrix $C$ is constructed using the squared Euclidean distance between points of the two distributions and scaled so that $\|C\|_{\max} = 1$. The total transported mass is then set to $s = 0.2 \min(\|x\|_{1}, \|y\|_{1})$. For all experiments, we fix the tolerance to $\texttt{tol} = 10^{-7}$, the regularization parameter to $\gamma = 10^{-3}$, and run ASPOT for 1500 iterations.  

\begin{figure}[t]
  \centering
  \includegraphics[width=0.85\linewidth]{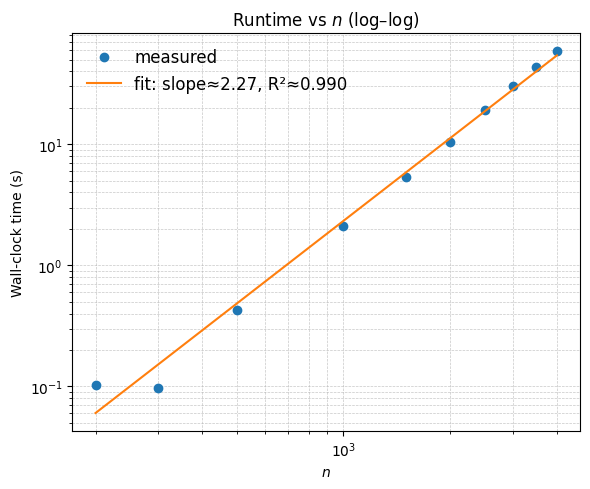}
  \caption{Scalability of ASPOT: runtime vs.~$n$ (log–log)}
  \label{wall_clock}
\end{figure}

We benchmark ASPOT over different problem sizes with $n \in \{200, 300, 500, 1000, 1500, 2000, 2500, 3000, 3500, 4000\}$. For each value of $n$, we record the runtime required for convergence. We then plot the results in a log--log scale, showing both the measured runtime and a linear regression line that fits the data (see Fig.~\ref{wall_clock}). The fitted slope is approximately $2.27$, which is consistent with the theoretical complexity of ASPOT, $\mathcal{O}(n^{7/3} \varepsilon^{-5/3})$.  

These results provide empirical evidence that the runtime of ASPOT scales as predicted, further validating our complexity analysis in the main text.

\textbf{Tuned Sinkhorn Technique.} An important observation from Fig.~\ref{fig:tunesink} is that the convergence curves for different values of $p$ (and thus different $\gamma$ choices in Algorithm~\ref{alg:tuned-sinkhorn-pot}) do not start at the same point. This difference arises directly from the initialization of the transport plan. Recall that in Algorithm~\ref{alg:tuned-sinkhorn-pot}, the kernel is defined as $\widetilde{K} = \exp(-\widetilde{C}/\gamma)$ and the transport matrix at each step is constructed as $X =\mathrm{diag}(e^{u})\,\widetilde{K}\,\mathrm{diag}(e^{v})$. Since $\gamma$ is tuned differently for each run, the kernel $\widetilde{K}$ changes, which in turn changes the initial $X$. Consequently, $\langle C,X\rangle$ differs at initialization for each $\gamma$, so the curves start at different points though they later follow similar convergence trends.

\textbf{Point Cloud Registration.} We briefly explain how point cloud registration can be formulated through partial optimal transport (POT), following \cite{https://doi.org/10.1111/cgf.14614}.  

Let two point clouds be $P=\{x_i\}_{i=1}^m \subset \mathbb{R}^3$ and $Q=\{y_j\}_{j=1}^n \subset \mathbb{R}^3$.  
We associate them with discrete distributions
\[
\mu = \sum_{i=1}^m \alpha_i \delta_{x_i}, 
\qquad 
\nu = \sum_{j=1}^n \phi_j \delta_{y_j},
\]
where $\alpha \in \mathbb{R}^m_+$, $\phi \in \mathbb{R}^n_+$, and $\sum_i \alpha_i=\sum_j \phi_j=1$.  

A transport plan $\pi \in \mathbb{R}^{m\times n}_+$ assigns partial mass between points of $P$ and $Q$.

With a rigid transformation $F(y)=Ry+t$ (rotation $R\in SO(3)$, translation $t\in \mathbb{R}^3$), the registration problem is
\[
\min_{R,t,\pi} \ \sum_{i=1}^m \sum_{j=1}^n \pi_{ij}\, \|x_i - (Ry_j+t)\|^2 ,
\quad \pi \in \mathcal{U}(r,c,s),
\]
where $\mathcal{U}(r,c,s)$ is defined in Section ~\ref{sec:preliminaries}.  

Given the optimal $\pi$, the rigid transformation is recovered by aligning weighted barycenters. Let
\[
u_x=\frac{1}{\|\pi\|_1}\sum_i \Big(\sum_j \pi_{ij}\Big) x_i,
\qquad
u_y=\frac{1}{\|\pi\|_1}\sum_j \Big(\sum_i \pi_{ij}\Big) y_j .
\]
We center the point clouds as $\hat{x}_i=x_i-u_x$, $\hat{y}_j=y_j-u_y$, and compute the singular value decomposition
\[
\hat{X}^\top \pi \hat{Y} = U \Sigma V^\top.
\]
The optimal rotation and translation are then
\[
R = V \operatorname{diag}(1,1,\det(VU^\top))U^\top, 
\qquad 
t = u_x - Ru_y.
\]

This setup connects POT with rigid alignment such that the transport plan captures soft correspondences, while the SVD step recovers the rigid motion.

\begin{figure}[t]
  \centering
  \includegraphics[width=1.05\linewidth]{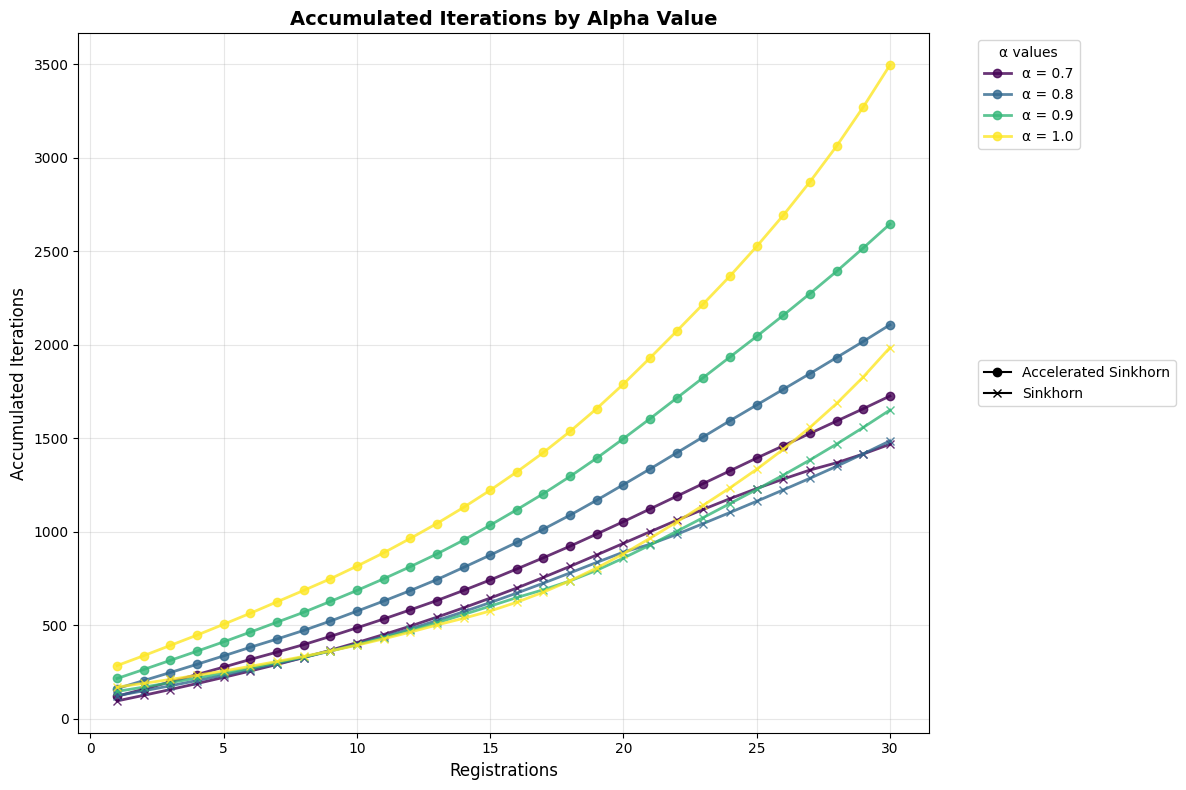}
  \caption{Accumulated iterations for larger $\alpha \in \{0.7,0.8,0.9,1.0\}$.}
  \label{fig:accum-largealpha}
\end{figure}

\begin{figure}[t]
  \centering
  \includegraphics[width=1.05\linewidth]{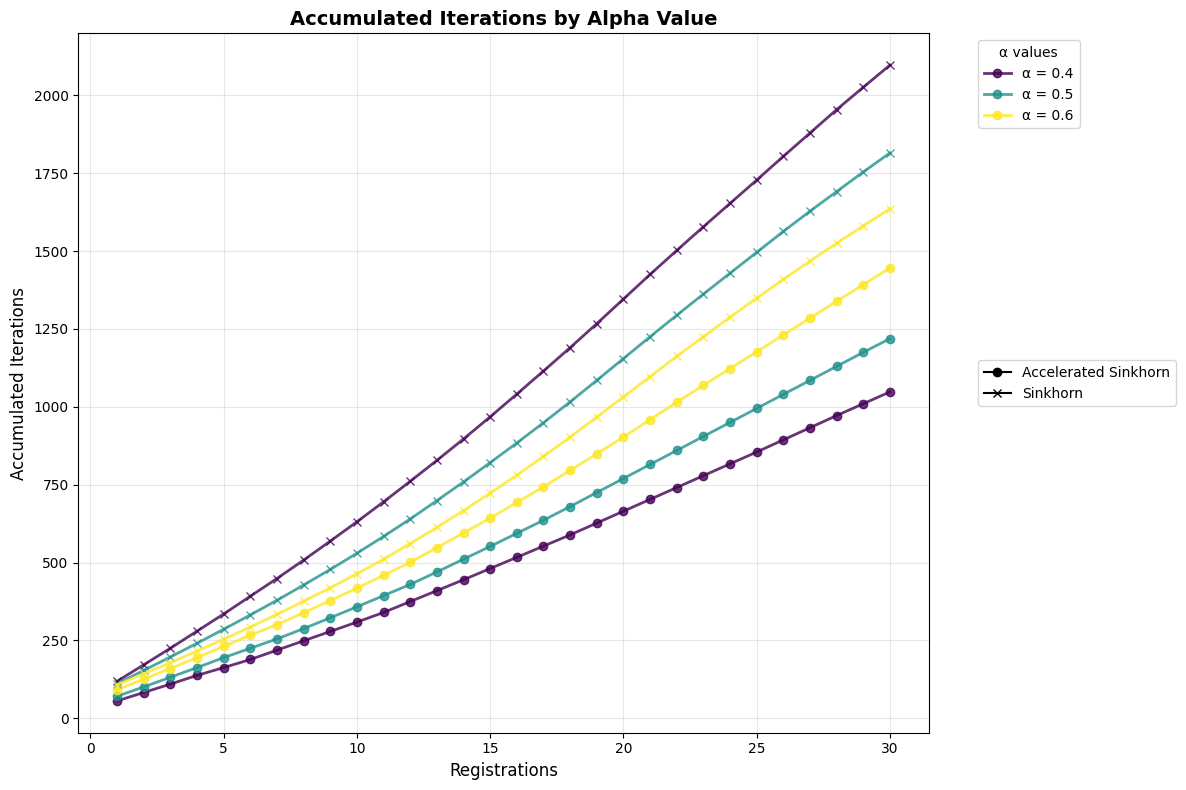}
  \caption{Accumulated iterations for small $\alpha \in \{0.4,0.5,0.6\}$.}
  \label{fig:accum-smallalpha}
\end{figure}

\textbf{Discussion.} Each curve in Fig. \ref{fig:accum-largealpha}, \ref{fig:accum-smallalpha} plots the accumulated number of inner iterations spent to compute POT maps as registration proceeds in Point Cloud Registration setting.  
The $x$–axis is the registration index, while the $y$–axis sums the iterations used up to that point.  
Thus, the slope of a curve directly reflects the per–registration iteration cost.  Across both panels, smaller $\alpha$ systematically lowers the slope, meaning each registration requires fewer inner iterations.  
This makes sense when a smaller transported mass reduces the active support of the plan, simplifying the subproblem and speeding up computation.  
For \textbf{larger} masses ($\alpha \ge 0.7$, Fig.~\ref{fig:accum-largealpha}), the classical Sinkhorn curves fall below ASPOT, showing that Sinkhorn is cheaper per registration in this high-mass regime. As $\alpha$ approaches $1.0$, this gap becomes even more pronounced, emphasizing that acceleration helps less when nearly the full mass must be transported.  For \textbf{smaller} masses ($\alpha \le 0.6$, Fig.~\ref{fig:accum-smallalpha}), the ordering reverses: ASPOT achieves a consistently lower slope and stays below Sinkhorn throughout, giving clear gains in the partial-transport setting that is most relevant for robust matching problems.  

When $\alpha$ is pushed very low ($\alpha \lesssim 0.3$), ASPOT still runs, but the outer registration may no longer perfectly align the point clouds. On the other hand, Sinkhorn often suffers from numerical instability at these values (divergence or NaNs), and its curves cannot even be completed. In practice, this trade-off leads to the choices in Section \ref{sec:pointcloud} we set $\alpha=0.4$ for ASPOT as it balances iteration cost with reliable alignment, and $\alpha=0.7$ for Sinkhorn.

\subsection{Domain Adaptation}

\textbf{Experimental Setup} Following the domain adaptation framework of \cite{courty2016optimaltransportdomainadaptation}, we evaluate POT methods on the two moons binary classification benchmark. In our setting, the source and target datasets contain different numbers of data points, with $N_s = 500$ samples in the source domain and $N_t = 600$ samples in the target domain. Both domains are generated from the two moons distribution with Gaussian noise level $\sigma = 0.05$ and centered at coordinates $(-0.5, -0.25)$. To simulate covariate shift, the target domain undergoes a rotation of $\theta = 60$ degrees while the source domain remains in its original orientation. To handle the unbalanced transport problem arising from $N_s \neq N_t$, we employ K-means clustering with $K = 500$ clusters on the target domain, replacing the 600 target samples with their cluster centroids where each centroid's weight corresponds to its cluster size. We denote ${r} \in \mathbb{R}^{500}$ as the source marginal distribution where $r_i = 1$ for all $i$, and ${c} \in \mathbb{R}^{500}$ as the target marginal distribution where $c_j$ equals the number of samples in cluster $j$. These marginals are normalized by dividing both ${r}$ and ${c}$ by $\max(\|{r}\|_1, \|{c}\|_1)$, yielding $\|{r}\|_1 = 0.75$ and $\|\mathbf{c}\|_1 = 1.0$. The cost matrix ${C} \in \mathbb{R}^{500 \times 500}$ is computed using squared Euclidean distances between source samples and target centroids, then normalized by its maximum value. For the partial transport formulation, we set the mass parameter $s = 0.999 \times \min(\|{r}\|_1, \|{c}\|_1)$. We compare three algorithms: (1) ASPOT with momentum-based acceleration, (2) APDAGD with our ROUND-POT procedure, and (3) standard feasible Sinkhorn. All algorithms use convergence tolerance $\varepsilon = 10^{-2}$ and maximum 3000 iterations. Once the transport plan ${T} \in \mathbb{R}^{500 \times 500}$ is computed, we apply barycentric mapping $\hat{{X}}_s = N_s({T} \times {X}_t)$ to transform source samples, then train a support vector machine with RBF kernel ($\sigma^2 = 1$) on $\{\hat{{X}}_s, {y}_s\}$ and evaluate on 20,000 test samples from the target distribution.

\textbf{Experimental Results}

\begin{figure}[t]
  \centering
  \includegraphics[width=1.05\linewidth]{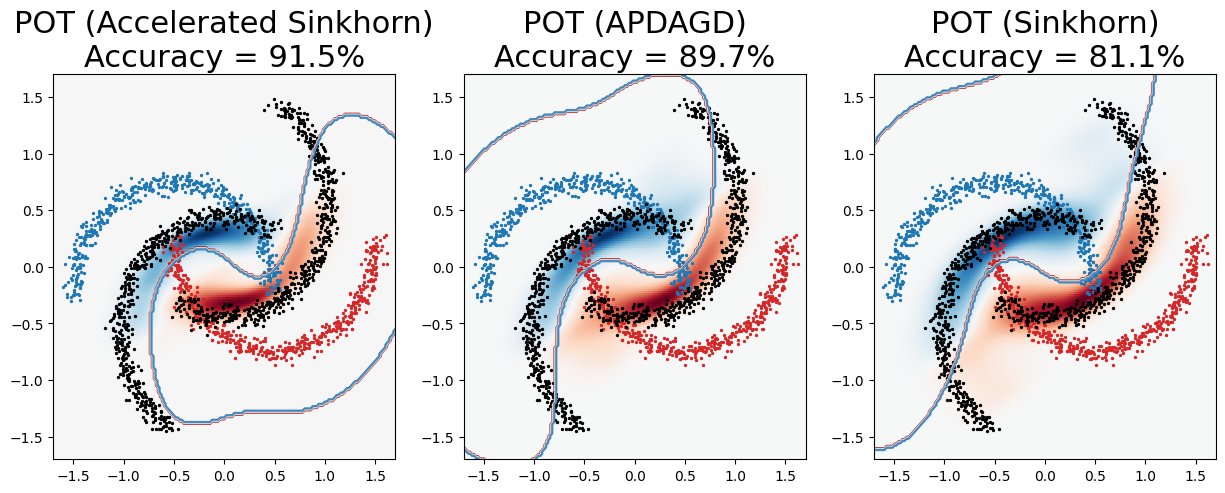}
  \caption{Domain adaptation on {two moons} with POT experiment.}
  \label{fig:pot-moons}
\end{figure}

Fig. \ref{fig:pot-moons} demonstrates that POT with ASPOT significantly outperforms the other methods, achieving $91.5\%$ accuracy on the target test set. This represents near-perfect domain adaptation, with the momentum-based acceleration enabling the algorithm to converge to superior transport plans that precisely align the source and target distributions. The decision boundary visualization reveals exceptionally smooth contours that closely follow the natural crescent shapes of the rotated two moons, with minimal misclassification regions and excellent class separation throughout the feature space. In comparison, POT with APDAGD achieves $89.7\%$ accuracy, falling $1.8$ percentage points short of ASPOT despite its exact mass constraint fulfillment through ROUND-POT. While APDAGD produces generally good decision boundaries, it exhibits more irregularities and some classification uncertainty in certain regions, suggesting that the rounding procedure's discrete optimization constraints lead to more conservative transport plans. POT with standard feasible Sinkhorn achieves the lowest accuracy at $81.1\%$, with its lack of acceleration causing slower convergence to suboptimal solutions that fail to fully capture the geometric correspondence between domains.

These results clearly establish ASPOT as the superior method for this domain adaptation task, demonstrating that momentum-based acceleration techniques are crucial for achieving high-quality transport plans in POT formulations. The $91.5\%$ accuracy achieved by ASPOT, compared to $89.7\%$ for APDAGD and $81.1\%$ for standard Sinkhorn, highlights the practical importance of algorithmic design choices beyond theoretical guarantees.

\section{Extra Algorithm}
We present a Sinkhorn algorithm with the tuning technique described in
Section~\ref{sec:color}. The method first converts POT to a balanced OT
problem via dummy nodes, then applies Sinkhorn with a principled choice of
\(\gamma\), and finally rounds the result back to a feasible POT plan.

\textbf{Sinkhorn Algorithm with tuning $\gamma$ technique}

\begin{algorithm}[H]
\caption{\textbf{Tuned Sinkhorn for POT}}
\label{alg:tuned-sinkhorn-pot}
\begin{algorithmic}[1]
\Require Marginals $r,c\!\in\!\R^n_+$, cost $C\!\in\!\R^{n\times n}_+$, budget $s$, accuracy $\varepsilon\!>\!0$, exponent $p\!\in\![1,\infty)$
\State $H_{\min}\!\gets\!\min\{H(r),H(c)\}$,\quad $\gamma\!\gets\!\big(2\varepsilon/(49H_{\min})\big)^{1/p}$,\quad $\varepsilon'\!\gets\!H_{\min}\,\gamma^{p}$
\State $A\!\gets\!8\|C\|_{\infty}/\varepsilon$
\State $\widetilde{C}\!\gets\!\begin{pmatrix} C & \mathbf{0}_n \\ \mathbf{0}_n^\top & A \end{pmatrix}\in\R^{(n+1)\times(n+1)}$
\State $\tilde r\!\gets\!\binom{r}{\|c\|_1-s},\quad \tilde c\!\gets\!\binom{c}{\|r\|_1-s}$
\State $\widetilde{K}\!\gets\!\exp(-\widetilde{C}/\gamma)$,\quad $u\!\gets\!\mathbf{0}_{n+1},\; v\!\gets\!\mathbf{0}_{n+1}$
\Function{B}{$u,v$} \Return $\mathrm{diag}(e^{u})\,\widetilde{K}\,\mathrm{diag}(e^{v})$ \EndFunction
\Repeat
  \State $u \gets u + \log \tilde r - \log\!\big(\mathrm{B}(u,v)\mathbf{1}_{n+1}\big)$
  \State $v \gets v + \log \tilde c - \log\!\big(\mathrm{B}(u,v)^\top \mathbf{1}_{n+1}\big)$
\Until{$\big\|\mathrm{B}(u,v)\mathbf{1}_{n+1}-\tilde r\big\|_1 + \big\|\mathrm{B}(u,v)^\top\mathbf{1}_{n+1}-\tilde c\big\|_1 \le \varepsilon'$}
\State $\widetilde{X}\!\gets\!\mathrm{B}(u,v)$
\State $\widetilde{X}\gets \textsc{Round}\big(\widetilde{X},\,\mathcal{U}(\tilde r,\tilde c)\big)$ \cite{altschuler2017near} 
\State \textbf{Output: } $\bar X \gets \widetilde{X}[1\!:\!n,\,1\!:\!n]$
\end{algorithmic}
\end{algorithm}


\end{document}